\PassOptionsToPackage{table}{xcolor}
\documentclass{article} 
\usepackage{2026_conference,times}

\usepackage{amsmath,amsfonts,bm}

\def\eqref#1{equation~\ref{#1}}

\def\1{\bm{1}}

\DeclareMathAlphabet{\mathsfit}{\encodingdefault}{\sfdefault}{m}{sl}
\SetMathAlphabet{\mathsfit}{bold}{\encodingdefault}{\sfdefault}{bx}{n}

\usepackage{hyperref}
\usepackage{url}
\usepackage{booktabs}
\usepackage{multirow}
\usepackage{enumitem}
\usepackage{graphicx}
\usepackage{wrapfig}
\usepackage{makecell} 
\usepackage{amssymb}
\usepackage{wasysym}
\usepackage{comment}
\usepackage{subcaption}
\usepackage{booktabs,tabularx,makecell}

\definecolor{iccvblue}{rgb}{0.21,0.49,0.74}
\hypersetup{
  colorlinks,
  citecolor=iccvblue,
  linkcolor=blue,
  urlcolor=blue}

\title{Macro-from-Micro Planning for High-Quality and Parallelized Autoregressive Long Video Generation}

\author{\textbf{Xunzhi Xiang}$^1$$^,$$^2$\textsuperscript{$*$}~~
    \textbf{Yabo Chen}$^2$$^,$$^3$\textsuperscript{$*$}~~
    \textbf{Guiyu Zhang}$^2$$^,$$^4$\textsuperscript{$*$}~~
    \textbf{Zhongyu Wang}$^2$\\
    \textbf{Zhe Gao}$^1$~~
    \textbf{Quanming Xiang}$^5$~~ 
    \textbf{Gonghu Shang}$^2$$^,$$^3$~~
    \textbf{Junqi Liu}$^2$~~
    \textbf{Haibin Huang}$^2$\\
    \textbf{Yang Gao}$^1$~~
    \textbf{Chi Zhang}$^2$~~
    \textbf{Qi Fan}$^1$\textsuperscript{$\dagger$}~~
    \textbf{Xuelong Li}$^2$\textsuperscript{$\dagger$}~~\\
    $^1$Nanjing University
    \quad
    $^2$Institute of Artificial Intelligence, China Telecom (TeleAI)\\
    $^3$Shanghai Jiao Tong University
    \quad
    $^4$Chinese University of Hong Kong, Shenzhen\\
    $^5$University of Chinese Academy of Sciences\\
}

\arxivfinalcopy 
\begin{document}

\maketitle
\vspace{-2em}
\begin{figure}[h]
    \centering
    \includegraphics[width=\textwidth]{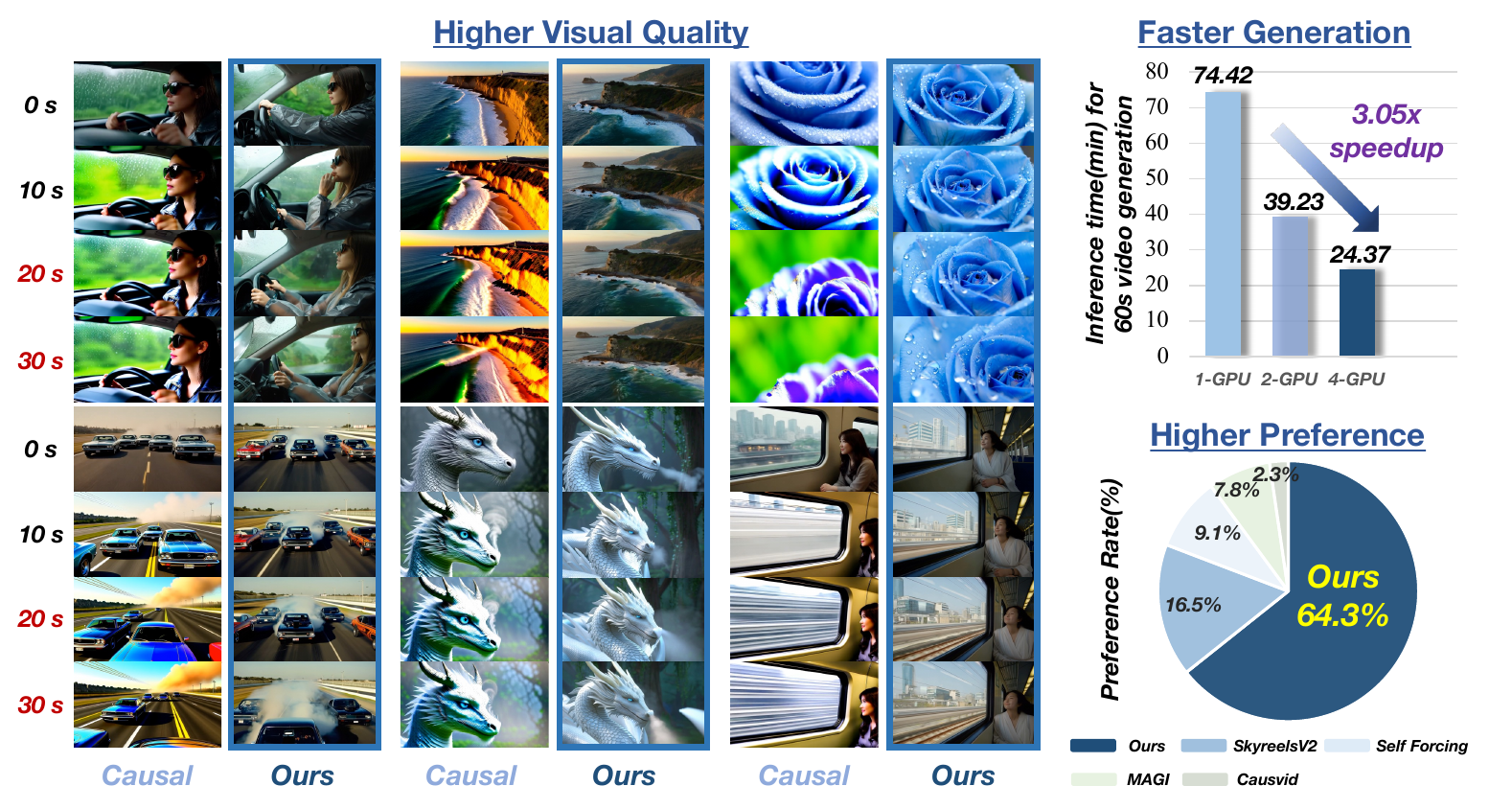}
    \vspace{-0.25in}
    \caption{We propose \textit{Macro-from-Micro Planning} (MMPL), a paradigm for long-video generation that achieves higher visual quality, faster speed, and stronger user preference than existing methods. Snapshots at 0s, 10s, 20s, and 30s (left) show robustness against temporal drift—semantic shifts, color changes, and structural artifacts—while quantitative results highlight accelerated multi-GPU inference (top-right) and dominant user preference (bottom-right).}
    \label{fig:demo_begin}
\end{figure}

\begin{abstract}
Current autoregressive diffusion models excel at video generation but are generally limited to short temporal durations.
Our theoretical analysis indicates that the autoregressive modeling typically suffers from temporal drift caused by error accumulation and hinders parallelization in long video synthesis. 
To address these limitations, we propose a novel planning-then-populating framework centered on Macro-from-Micro Planning (MMPL) for long video generation. MMPL sketches a global storyline for the entire video through two hierarchical stages: {\textit{Micro Planning and Macro Planning}}. Specifically, \textit{Micro Planning} predicts a sparse set of future keyframes within each short video segment, offering motion and appearance priors to guide high-quality video segment generation.
\textit{Macro Planning} extends the in-segment keyframes planning across the entire video through an autoregressive chain of micro plans, ensuring long-term consistency across video segments. 
Subsequently, MMPL-based \textit{Content Populating} generates all intermediate frames in parallel across segments, enabling efficient parallelization of autoregressive generation.
The parallelization is further optimized by \textit{Adaptive Workload Scheduling} for balanced GPU execution and accelerated autoregressive video generation.
Extensive experiments confirm that our method outperforms existing long video generation models in quality and stability. 
Generated videos and comparison results are in the \href{https://nju-xunzhixiang.github.io/Anchor-Forcing-Page/}{Demo page}.
\end{abstract}

\section{Introduction}
Long video generation is crucial for applications such as movie production \citep{MovieGen, MovieDreamer}, virtual reality \citep{CAT4D, WorldV}, and digital human creation \citep{Animateanyone, ReMask, Proteus, champ}. Despite significant advances in video synthesis, creating extended sequences with both temporal coherence and computational efficiency remains challenging \citep{exposure-1}. 

Conventional diffusion-based methods \citep{DiT,Wan,GenTron,videocrafter,WALT,Latte} have achieved remarkable quality by jointly optimizing all frames via bidirectional attention. However, this global optimization necessitates the simultaneous generation of the entire sequence, introducing significant latency and rendering these methods impractical for real-time or interactive scenarios.

Autoregressive (AR) models \citep{PAR,randar,zipar} offer an effective alternative by sequentially generating images or frames. This incremental strategy enables users to start viewing immediately after the initial frames are available, greatly reducing latency. Furthermore, AR models impose fewer constraints on video duration and facilitate interactive user control. Representative AR methods such as VideoGPT \citep{VideoGPT}, LBD \citep{LBD}, and CogVideo \citep{Cogvideo} adopt a next-frame prediction paradigm based on discrete tokenizers, substantially lowering latency compared to diffusion-based approaches. However, their reliance on discrete tokenization inherently leads to quantization artifacts, reducing visual fidelity.
Hybrid AR-diffusion methods \citep{ardiffusion,diffusionforcing,historydiffusion} merge autoregressive generation with continuous diffusion processes to overcome these limitations. By integrating diffusion into the autoregressive framework, these methods avoid discrete codebooks, effectively addressing quantization-induced degradation and significantly improving output quality.

Nevertheless, both AR and AR-diffusion methods suffer from error accumulation. Since each frame depends explicitly on previously generated frames, errors from early frames compound and magnify over subsequent predictions, causing long-term degradation and temporary drift. Moreover, existing autoregressive approaches remain strictly sequential, inherently preventing parallel generation and thus limiting computational efficiency and scalability. These fundamental challenges motivate the question: 
\textit{How can AR models move beyond naive autoregressive modeling to enable high-quality and parallelized long-video synthesis?}

Analogous to the workflow of professional filmmakers, long video creation naturally benefits from a hierarchical \textit{plan-then-populate} paradigm. In a typical movie production, the process does not proceed by shooting every frame in chronological order. Instead, the production team first develops a Macro Plan, a rough storyboard that captures the overall structure and key moments of the film. This Macro Plan consists of multiple Micro Plans, each representing an individual scene or shot. With this setup, different scenes can be filmed in parallel according to their Micro Plans, much like multiple crews shooting on separate sets at the same time. The Macro Plan then coordinates and assembles all these pieces into a coherent long movie. Such hierarchical planning improves the efficiency of film production while ensuring that the final movie remains seamless and coherent.
Building on this insight, we first perform a systematic analysis of error accumulation in AR and Non-AR video generation, revealing the fundamental mechanisms that drive long-term drift.
Guided by these findings, we propose a novel plan-then-populate framework centered on Macro-from-Micro Planning (MMPL) for scalable, high-quality long video generation. 
MMPL operates via two complementary planning levels: \textit{Micro Planning} efficiently predicts multiple keyframes of each segment simultaneously from its initial frame, capturing detailed local trajectories; \textit{Macro Planning} autoregressively chains these segments by initializing each segment $S$ from the last keyframe of segment $S-1$, thus ensuring global narrative coherence across the entire video.
Once all keyframes are established, MMPL-based Content Populating concurrently synthesizes intermediate frames between keyframes within each segment, adhering to boundary constraints and eliminating sequential frame dependencies. To further optimize pipeline efficiency, we introduce an adaptive workload scheduling strategy that dynamically allocates GPU resources. This approach significantly reduces the overall generation time to approximately one-third of the original, without relying on distillation-based acceleration, while preserving high visual fidelity.

Overall, our work delivers the following contributions:
\begin{itemize}[leftmargin=2em, itemsep=0pt, parsep=0pt]
    \item We propose \textit{Macro-from-Micro}, a hierarchical autoregressive planning method that forms coherent global storylines across segments of the entire video, while drastically reducing temporal error accumulation in long-video generation.
    \item We propose MMPL-based Content Populating, which synthesizes frames for multiple segments in parallel under the guidance of pre-planned keyframes, breaking the intrinsic sequential bottleneck of conventional autoregressive pipelines.
    \item We further design an adaptive multi-GPU workload scheduling strategy that balances segment generation across devices, substantially reducing wall-clock time for long-video synthesis.
\end{itemize}

\begin{figure*}[t]
    \centering
    \includegraphics[width=0.95\textwidth]{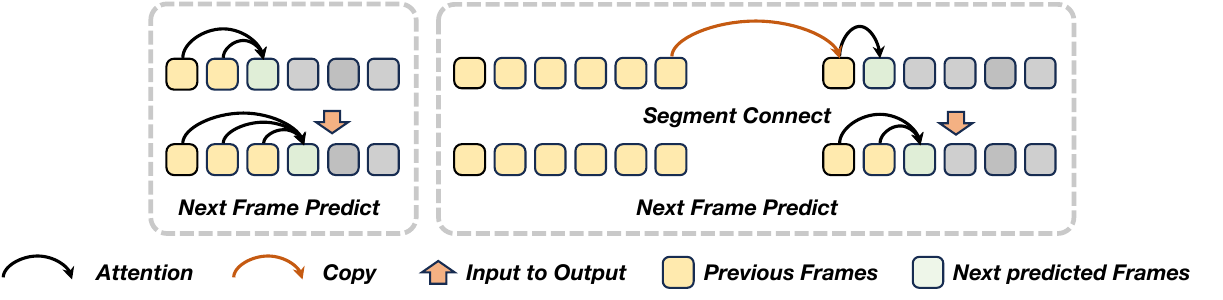}
    \vspace{-0.05in}
    \caption{Existing AR methods generate frames sequentially in a step-by-step manner,
    inevitably causing error accumulation (as shown in Figure~\ref{fig:demo_begin}) and prohibiting parallel generation.}
    \label{fig:compare}
    \vspace{-0.1in}
\end{figure*}

\section{Related Work}

\textbf{Bidirectional Diffusion Models for Video Generation.}
Diffusion models have emerged as a dominant approach for high-quality visual synthesis, benefiting from their scalability and superior generative capabilities \citep{LDM, beatsgan}. In video generation, existing diffusion architectures primarily rely on bidirectional attention mechanisms to jointly denoise all frames within a sequence \citep{animatediff, videodiffusionmodels, alignlatent, IM-Zero, zhang2024vast}. While this enables high-fidelity outputs, the requirement to concurrently generate entire sequences prohibits streaming or incremental video generation, resulting in significant inference latency and hindering applications involving long video generation.

\textbf{Causal Autoregressive Models for Video Generation.}
Autoregressive (AR) models provide an alternative by sequentially generating video frames or spatiotemporal tokens, conditioning each new frame on previously generated content \citep{ardiffusion,diffusionforcing,historydiffusion,NOVA,vidgpt,ARLON}. This causal generation paradigm naturally supports streaming outputs and substantially reduces initial latency. However, the sequential dependency between frames inherently introduces error accumulation. As prediction chains grow longer, these errors compound, resulting in temporal drift and degraded visual coherence, especially noticeable in extended video sequences (Figure~\ref{fig:compare}).

\textbf{Methods for Long Video Generation.}
Long video synthesis poses unique challenges due to cumulative errors and computational bottlenecks inherent in autoregressive inference. Recent efforts, such as CausVid \citep{Causvid} and Self Forcing (SF) \citep{SelfForcing}, address these challenges by introducing methods like \textit{Diffusion Forcing} and \textit{Self Forcing}, aimed at reducing the mismatch between training and inference dynamics. Although these techniques partially alleviate drift through recursive conditioning and short-step diffusion, they remain susceptible to significant error propagation when generating videos exceeding approximately 30 seconds.

\textbf{Planning Prediction.} 
A closely related work, FramePack-Plan~\citep{zhang2025framepack}, mitigates error accumulation via step-wise frame jumping, and compresses context to extend video length. In contrast, our Macro-from-Micro framework introduces three key innovations. First, we adopt a two-level hierarchical planning scheme: a Micro Plan predicts segment-level keyframes, and a Macro Plan, composed of overlapping Micro Plans, forms a coherent global storyline through autoregressive scheduling. Second, each Micro Plan produces all pre-planned keyframes for its segment in a single forward pass conditioned only on the initial frame, drastically compressing the autoregressive chain. Finally, once the Macro Plan is obtained, the remaining content within all segments is synthesized in parallel, achieving high throughput while preserving temporal coherence.

\section{Method}
\subsection{Macro-from-Micro Planning}
\label{method_sec_1}
Motivated by the analysis in the supplementary material, we observe that 
autoregressive models accumulate errors proportionally to the number of propagation steps, 
whereas non-autoregressive models decouple errors from the step count through joint optimization. 
To exploit the complementary strengths of both paradigms, 
we introduce \textit{Macro-from-Micro Planning (MMPL)}, 
a unified planning method comprising two key components: 
\textit{Micro-Planning} and \textit{Macro-Planning}.
\label{train_method}

\begin{figure*}[t]
    \centering
    \includegraphics[width=0.95\textwidth]{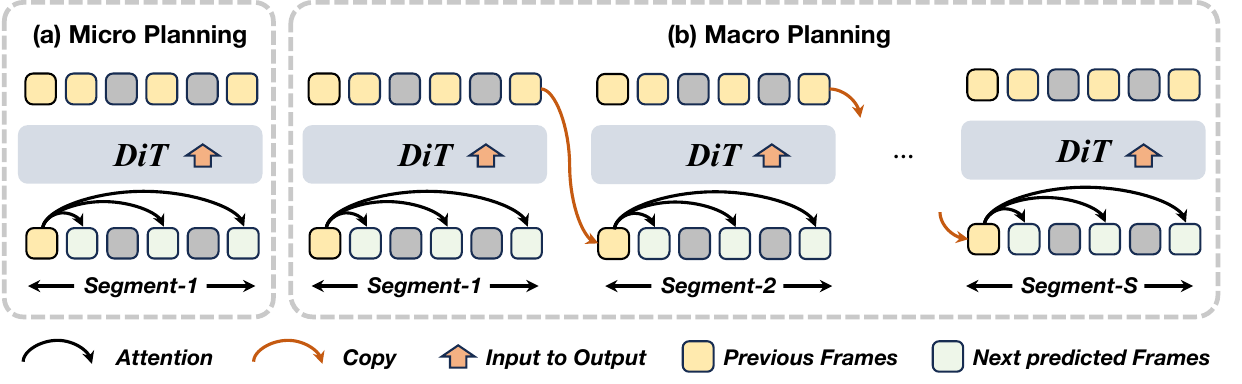}
    \vspace{-0.05in}
    \caption{Overall framework of Macro-from-Micro Planning. Our method operates on two planning levels: (1) Micro Planning, which predict a sequence of future frames within each segment to mitigate local error accumulation, and (2) Macro Planning, formed as an Autoregressive Chain of Micro Plans, where the planning frames of the first segment autoregressively generate the planning frames of subsequent segments, ensuring long-horizon temporal consistency.}
    \label{fig:ours}
    \vspace{-0.1in}
\end{figure*}

\textbf{Micro Planning.}
\label{Micro}
As illustrated in Figure~\ref{fig:ours},
Micro Planning $\mathcal{M}$ constructs a short temporal storyline for each segment
with N frames by predicting a small set of key frames that act as stable anchors for subsequent content synthesis.
This sparse set of \textit{pre-planning frames},
${x^{t_a}, x^{t_b}, x^{t_c}}$,
is jointly predicted from the initial frame $x^1$.
This process can be expressed as:
\begin{equation}
p(\mathcal{M} \mid x^1)
= p\big(x^{t_a}, x^{t_b}, x^{t_c} \mid x^1\big),
\quad
\mathcal{M} := { x^{t_a}, x^{t_b}, x^{t_c} }.
\end{equation}
Where $t_a = 2$ denotes the earyl neighbor of the initial frame, 
$t_b = N/2$ serves as the global midpoint, 
and $t_c = N$ marks the terminal frame of the sequence. 
These \textit{pre-planning frames} are jointly optimized while conditioned solely on the initial frame $x^1$, 
rendering their mutual drift with $x^1$ negligible. 
Moreover, since all frames are jointly optimized from the initial frame $x^1$, their residual errors are mutually constrained and remain negligible, preventing the cumulative drift characteristic of sequential autoregressive generation.
This design ensures temporal coherence within each segment
and establishes a stable, drift-resistant foundation
for the subsequent populating process.

\textbf{Macro Planning.}
\label{Macro}
While Micro Planning provides a segment-level temporal storyline,
it remains limited in capturing global dependencies across the entire video.
To achieve long-range coherence,
we extend Micro Planning into \textit{Macro Planning}, denoted as $\mathcal{M}^{+}$.
Macro Planning constructs a global storyline for the entire long video
by sequentially chaining overlapping Micro Plannings across video segments.
Concretely, the terminal pre-planning frames of one segment
serve as the initial conditions for the next,
thereby linking local plans into a coherent long-horizon structure,
which can be regarded as a segment-level autoregressive process over the video timeline.
Let the full video of frame length $T$ be partitioned into $S$ short segments,
with the first frame of the $s$-th segment denoted as $x_s^1$.
This process can be expressed as:
\begin{equation}
p(\mathcal{M}^{+} \mid x^1)
= \prod_{s=1}^{S} p(\mathcal{M}_s \mid x_s^1),
\quad
\mathcal{M}^{+} := \bigcup_{s=1}^{S} \mathcal{M}_s ,
\end{equation}
where $\mathcal{M}_s$ represents the Micro Planning for the $s$-th segment.
By hierarchically chaining these segment-level plans, Macro Planning transforms the original frame-by-frame long-range autoregressive dependency into a segment-wise sequence of sparse planning dependencies. This restructuring preserves global temporal coherence by ensuring a consistent storyline across segments and suppresses temporary drift, effectively reducing the error accumulation scale from the $T$-frame level of conventional autoregressive generation to the $S$-segment level under our framework, where $S \ll T$.

\begin{wrapfigure}{r}{0.21\textwidth}
\vspace{-1.6em}
    \centering 
    \includegraphics[width=0.21\textwidth]{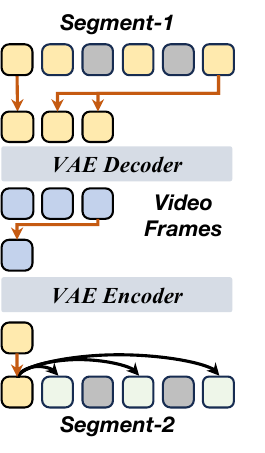}
    \vspace{-2em}
    \caption{Our Re-Encoding and Decoding Strategy.}
    \label{fig:vae}
\vspace{-0.8em}
\end{wrapfigure}

However, when linking Micro Plannings through an autoregressive chain, directly reusing the tail latent tokens of the preceding segment as the prefix for the next often leads to boundary flickering and color shifts across segment transitions. This issue stems from a distribution mismatch. The first latent frame fundamentally differs from the others: it represents only the initial image, while subsequent frames incorporate temporally compressed information, resulting in inconsistent statistics across frames. Therefore, inspired by CausVid \citep{Causvid}, we introduce a drift-resilient re-encoding and decoding strategy to stabilize inter-segment transitions.
Specifically, as shown
in Figure~\ref{fig:vae}, we first concatenate the initial latent token of the preceding segment with its terminal planning tokens and feed the sequence into the VAE decoder for video reconstruction.
However, since VAE decoding requires each token to condition on strictly contiguous temporal prefixes, any temporal discontinuity in the input sequence leads to pronounced color shifts and boundary artifacts. To mitigate this issue, we duplicate the terminal planning tokens once and insert the copy between the initial latent token and the original terminal planning tokens, forming a temporally contiguous latent sequence for decoding.
After reconstruction, we re-encode the second copy of the terminal planning tokens and use the resulting latents as the initial tokens for the next segment’s Micro Planning.
This design enforces both statistical and temporal consistency in the latent space, effectively suppressing color shifts and boundary flickering, and achieving smooth, stable inter-segment transitions.

\subsection{MMPL-based Content Populating}

\begin{wrapfigure}{r}{0.22\textwidth}
\vspace{-1em}
    \centering 
    \includegraphics[width=0.22\textwidth]{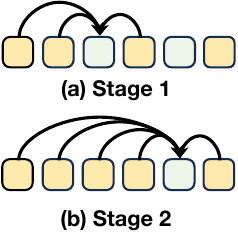}
    \caption{Two Stages of our MMPL-based Content Populating.}
    \label{fig:populate}
\vspace{-1.4em}
\end{wrapfigure}
\label{Populating}
Following Sec.~\ref{method_sec_1}, the Micro Plan $\mathcal{M}$ naturally partitions each video segment into two 
\textit{sub-segments}, bounded by consecutive planning frames, e.g., 
$\big[x^{t_a}, x^{t_b}\big]$ and $\big[x^{t_b}, x^{t_c}\big]$.  
To synthesize the complete segment by populating the remaining frames under the constraints of these planning frames,  
we introduce {MMPL-based Content Populating}.  
Specifically, Micro Planning generates three types of planning frames: 
\textit{early}, \textit{midpoint}, and \textit{terminal}. 
Inspired by early methods that generate videos conditioned on the first and last frames, 
we divide the Content Populating process into two stages, as shown in Figure~\ref{fig:populate}. 
In the first stage, we populate the interval by using the initial and early planning frames as the head and the midpoint planning frames as the tail, synthesizing the intermediate content.
In the second stage, we extend the populated sequence by taking all frames between the initial frame and the midpoint planning frames as the new head and the terminal frames as the tail, thereby generating the remaining content. This process can be expressed as:
\begin{equation}
p(\mathcal{C}_i \mid \mathcal{M}_i)
= p\big(x_i^{t_a+1:t_b-1} \mid x_i^{1:t_a}, x_i^{t_b}\big)
\cdot
p\big(x_i^{t_b+1:t_c-1} \mid x_i^{1:t_b}, x_i^{t_c}\big),
\end{equation} 
where $\mathcal{C}_i$ corresponds to the frames to be synthesized in the $i$-th segment.  
The variables $x_i^{t_a}$, $x_i^{t_b}$, and $x_i^{t_c}$  
denote the early, midpoint, and terminal planning frames of segment $i$, respectively.  
The notation $x_i^{1:t_a}$ and $x_i^{1:t_b}$ indicates that the generation of each sub-segment  
is conditioned not only on its boundary planning frames but also on all preceding frames in the same segment.  
Accordingly, the intermediate frames within the two sub-segments, denoted as $x_i^{t_a+1:t_b-1}$ and $x_i^{t_b+1:t_c-1}$, represent the remaining content to be populated.

This factorization explicitly demonstrates that content population within each sub-segment depends exclusively on its corresponding planning frames. Consequently, multiple sub-segments can be independently optimized in parallel, provided their internal planning frames have been generated.
Furthermore, leveraging multiple GPUs, the proposed MMPL-based Content Populating can distribute segment-wise optimization across different devices, enabling concurrent execution. This parallelization significantly enhances computational efficiency, facilitating highly efficient long-video synthesis. Formally, this parallel generation process can be expressed as:
\begin{equation}
p(\mathcal{C} \mid \mathcal{M})
= \prod_{i=1}^{S} p(\mathcal{C}_i \mid \mathcal{M}_i),
\end{equation}
where the global video synthesis task factorizes into independent segment-level sub-tasks, each executable in parallel across multiple GPUs once their planning frames have been fully established.

\begin{figure}[t]
    \centering
    \includegraphics[width=\textwidth]{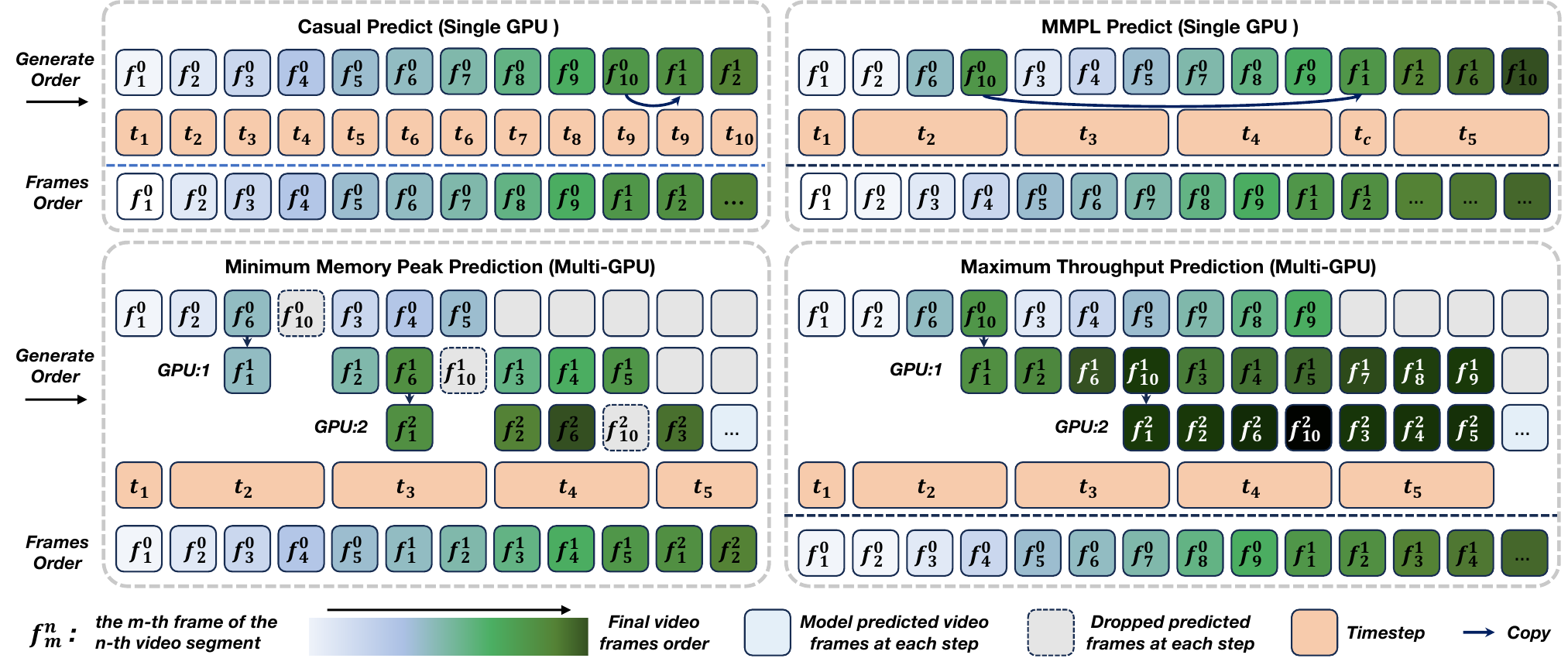}
    \vspace{-0.2in}
    \caption{
    Multi-GPU parallel inference via adaptive workload scheduling.  
    Given the initial frame $f^0_1$, segment 0 first generates its planning frames  
    $f^0_2$, $f^0_6$, and $f^0_{10}$.  
    These planning frames then guide the content population of the intermediate frames  
    $f^0_3$, $f^0_4$, and $f^0_5$.  
    While segment 0 is still populating these frames,  
    segment 1 can immediately start its Micro Planning  
    by taking $f^0_{10}$ as the initial frame $f^1_1$  
    and generating its own planning frames $f^1_2$, $f^1_6$, and $f^1_{10}$.  
    This staged execution enables overlapping planning and populating across segments,  
    maximizing multi-GPU parallelism.
    }
    \label{fig:parallel_inference_method}
    \vspace{-0.1in}
\end{figure}

\subsection{adaptive workload scheduling}

As discussed in Sec.~\ref{Populating}, the content populating of different segments can be executed in parallel across multiple GPUs.  
However, this approach suffers from a key limitation: parallelization cannot start until the planning frames of all segments have been fully generated, introducing an inevitable prefix delay that degrades the overall pipeline throughput.  
To further improve generation efficiency, we propose an \textit{adaptive workload scheduling} strategy, which dynamically adjusts the execution order of Micro Planning, Macro Planning, and Content Populating to maximize parallelism. Specifically, Macro Planning is constructed as an autoregressive chain of segment-level Micro Plannings, 
which naturally enforces a strict generation order of planning frames across segments. 
This property allows us to initiate the Content Populating of an earlier segment as soon as its planning frames are available, 
without waiting for the planning frames of all subsequent segments to finish. 
To illustrate the workload scheduling, consider a case where we set $t_a = 2$, $t_b = 6$, and $t_c = 10$ to evenly cover the temporal span.
As shown in Figure~\ref{fig:parallel_inference_method}, the planning frames of the current segment, generated via \textit{Micro Planning} ($x^{t_b}_s$ or $x^{t_c}_s$), immediately serve as the initial frame $x^1_{\text{s+1}}$ for the subsequent segment.
This allows the next segment to start its own \textit{Micro Planning} while the current segment is still performing \textit{Content Populating} to generate $x^{t_a+1:t_b-1}_{s}$.
This staged independence naturally enables segment-parallel generation, 
as formally expressed in Eq.~(\ref{eq:segment_parallel}):
\begin{equation}
\begin{aligned}
& \text{Segment s:} && x_{s}^{t_a+1:t_b-1} 
   \sim p_\theta(x \mid x_{s}^{1}, x_{s}^{t_a}, x_{s}^{t_b}), \\
& \text{Segment s+1:} && \{x_{s+1}^{t_a}, x_{s+1}^{t_b}, x_{s+1}^{t_c}\} 
   \sim p_\theta(x \mid x^1_{s+1}), 
   \quad x^1_{s+1} \in \{x_{s}^{t_b}, x_{s}^{t_c}\}.
\end{aligned}
\label{eq:segment_parallel}
\end{equation}
Here, the initial frame \(x^1_{s+1}\) of the next segment can be selected 
either as \(x_{s}^{t_b}\) or \(x_{s}^{t_c}\). 
This selection directly determines the parallel execution strategy 
and leads to two distinct modes:

\textbf{(1) Minimum Memory Peak Prediction.} 
When $x^{t_b}_s$ is used as $x^1_{\text{s+1}}$, 
intermediate frames $x^{t_b+1}:x^{t_c-1}$ are skipped, 
bypassing the region with the deepest temporal context 
and highest generation latency. 
This mode minimizes peak memory usage and reduces per-segment latency 
but introduces frame reuse between segments, 
slightly reducing overall throughput.

\textbf{(2) Maximum Throughput Prediction.} 
When $x^{t_c}_{s}$ is used as $x^1_{\text{s+1}}$, 
all intermediate frames are generated sequentially within the segment, 
eliminating inter-segment redundancy 
and achieving maximal pipeline efficiency, 
at the cost of higher per-segment computation.

These two execution strategies offer a trade-off between 
local memory/latency and global throughput, 
allowing flexible deployment choices.

\section{Experiments}

\textbf{Baselines.} 
We compare our model against representative open-source video generation systems of comparable scale, including FIFO~\citep{FIFO}, SkyReelsV2~\citep{SkyReels}, MAGI~\citep{MAGI}, CausVid~\citep{Causvid}, and Self Forcing~\citep{SelfForcing}. 
All methods are evaluated under a unified sliding-window protocol, where each fixed-length segment (e.g., 5\,s) is causally conditioned on the final frames of the preceding segment. 
We adopt SkyReels-V2-14B and MAGI-4.5B as our primary baselines, while CausVid and Self Forcing (1.3B, distilled from 14B teachers) serve as high-fidelity autoregressive representatives. 

\textbf{Training Details.} 
We implement \textit{MMPL} on Wan2.1-T2V-14B~\citep{Wan}, a DiT-based~\citep{DiT} Flow Matching model originally built for 5-second videos. 
To support long-horizon modeling, we adopt FlexAttention~\citep{flexattention} for scalable training and FlashAttention-v3~\citep{flashattention} for fast inference. 
The model is fine-tuned on 50k curated high-quality videos at $832{\times}480$ resolution, ensuring diverse and clean data for stable optimization. 
Training runs for 8{,}000 iterations on 32 H100 GPUs with AdamW at a $1\times10^{-5}$ learning rate. 
For hierarchical planning, we set $t_a = 2,3,4$, $t_b = 9,10,11$, and $t_c = 19,20,21$, corresponding to early, midpoint, and late planning frames guiding segment-wise generation. 
Additional hyperparameters and ablation settings are provided in the supplementary material.

\textbf{Evaluation.}  
We evaluate on the VBench-long benchmark \citep{vbench2}, which assesses subject and background consistency, motion smoothness, aesthetic quality, and imaging quality, jointly reflecting temporal stability and perceptual fidelity. 
For the main study, we generate 30s videos from 120 randomly sampled MovieGen \citep{MovieGen} prompts on a single H100 GPU. 
To complement these quantitative metrics, we also conduct a user study: for each baseline, 19 videos of about 30\,s are generated using the first 19 MovieGen prompts, and 29 participants perform pairwise comparisons, selecting the video that better matches the prompt in terms of visual quality and semantic fidelity. 
This combination of objective metrics and human judgments provides a rigorous evaluation of both numerical performance and perceptual quality. 
Details of the user study are provided in the supplementary material.

\begin{table*}[t]
\centering
\scriptsize
\setlength{\tabcolsep}{3pt}
\renewcommand{\arraystretch}{1}
\caption{Evaluation metrics for the other baselines and MMPL. The first five metrics are automatically computed by VBench,
while the last three are obtained through human evaluation.}
\begin{tabular}{l|ccccc|ccc}
\toprule
 & \multicolumn{5}{c|}{\textit{VBench Evaluation}} & \multicolumn{3}{c}{\textit{Human Evaluation}} \\
\cmidrule(lr){2-6} \cmidrule(lr){7-9}
\textbf{Model} & 
\textbf{Subject} & \textbf{Background} & \textbf{Motion} & \textbf{Aesthetic} & \textbf{Imaging} & 
\textbf{Text-Visual} & \textbf{Content} & \textbf{Color} \\
 & 
\textbf{Consistency} & \textbf{Consistency} & \textbf{Smoothness} & \textbf{Quality} & \textbf{Quality} & 
\textbf{Alignment} & \textbf{Consistency} & \textbf{Shift} \\
\midrule
\rowcolor{gray!10}
\multicolumn{9}{l}{\textit{Causal}} \\
\quad FIFO~\citep{FIFO}      & 0.956 & 0.960 & 0.949 & 0.588 & 0.603  &-  &-  &-  \\
\midrule
\rowcolor{gray!10}
\multicolumn{9}{l}{\textit{Distilled Causal}} \\
\quad CausVid(\citep{Causvid})             & 0.969 & \textbf{0.980} & 0.981 & \underline{0.606} & \underline{0.661}  &  34.7 & 33.0 & 25.0 \\
\quad SF~\citep{SelfForcing}        & 0.967 & 0.958 & 0.980 & 0.593 & \textbf{0.689}  & \underline{52.0}  & 46.1 & 50.5 \\
\midrule
\rowcolor{gray!10}
\multicolumn{9}{l}{\textit{DF Causal}} \\
\quad SkyReels~\citep{SkyReels}       & 0.956 & 0.966 & \underline{0.991} & 0.600 & 0.581  & 47.9 & \underline{51.4} & \underline{51.3} \\
\quad MAGI-1~\citep{MAGI}             & \underline{0.979} & \underline{0.970} & \underline{0.991} & 0.604 & 0.612  & 34.7 & 40.4 & 39.5 \\
\midrule
\rowcolor{gray!10}
\multicolumn{9}{l}{\textit{Planning}} \\
\quad \makecell[l]{MMPL} 
      & \textbf{0.980} & 0.968 & \textbf{0.992} & \textbf{0.628} & \underline{0.661} & \textbf{80.0} & \textbf{79.2}  & \textbf{83.1} \\
\bottomrule
\end{tabular}
    \vspace{-0.05in}
\label{tab:results_consistency_metrics}
    \vspace{-0.1in}

\end{table*}

\begin{figure*}[t]
    \centering
    \includegraphics[width=0.9\textwidth]{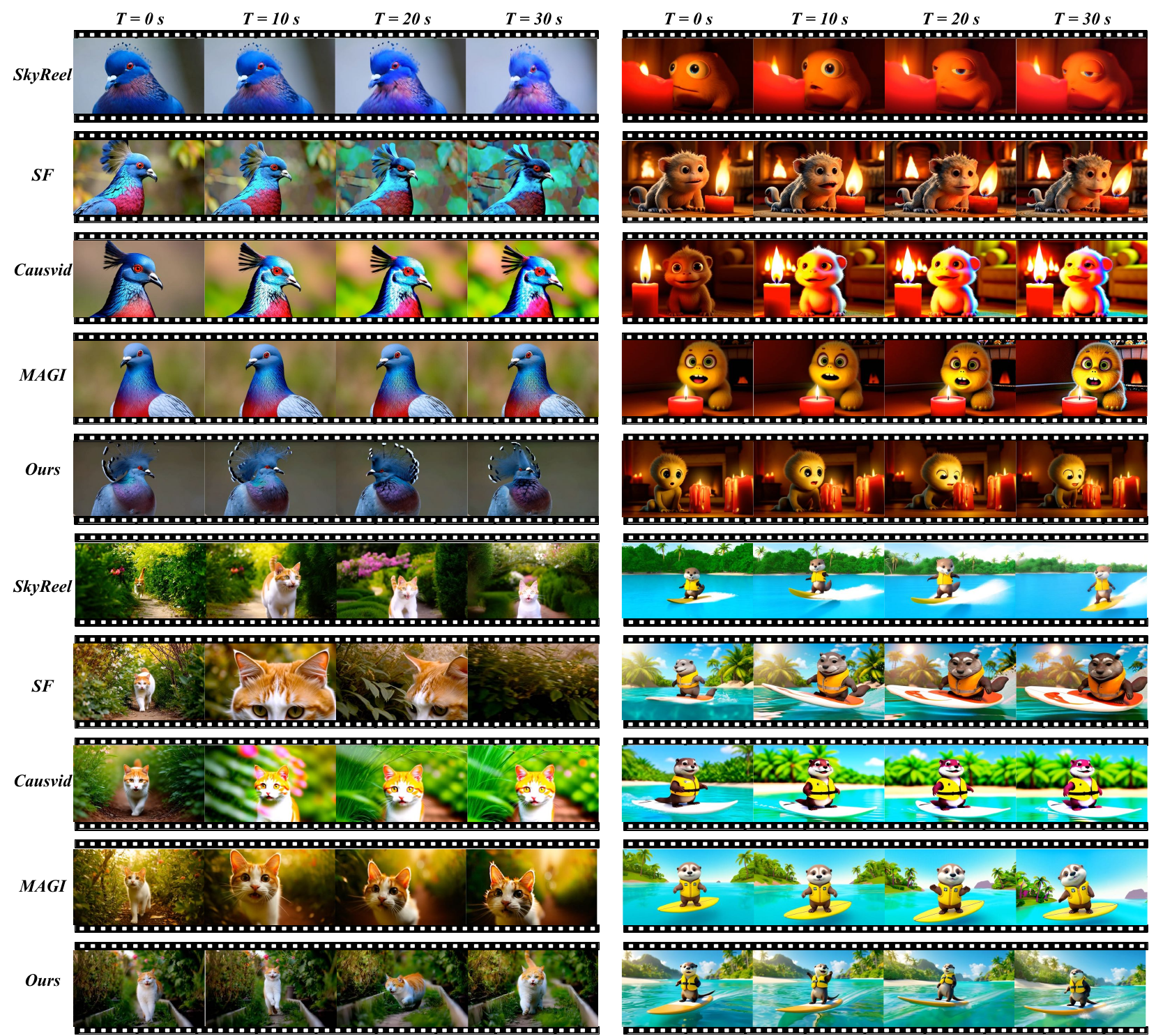}
    \caption{Qualitative comparisons. We visualize videos generated by Macro-from-Micro against those by MAGI, SkyReels-V2, Self Forcing, and CausVid. }
    \label{fig:Qualitative}
    \vspace{-0.1in}

\end{figure*}

\textbf{Quantitative Results.}
As shown in Table~\ref{tab:results_consistency_metrics}, our Macro-from-Micro method achieves the strongest overall performance on VBench, excelling in subject consistency 0.980, motion smoothness 0.992, and aesthetic quality 0.628, while maintaining competitive imaging quality 0.661 and only slightly lower background consistency 0.968 than CausVid and MAGI-1. However, VBench metrics, particularly subject and background consistency, tend to favor less dynamic scenes and cannot fully capture the perceptual complexity of long video generation. To address this limitation, we conducted a human study by generating 19 diverse 30-second videos per method, spanning humans, vehicles, and natural landscapes. Thirty participants rated each video on text-visual alignment, content consistency, and long-sequence color stability. Our method achieved the highest scores in all three dimensions: 80.0 for text-visual alignment, 79.2 for content consistency, and 83.1 for color stability, substantially outperforming other baselines. Besides, as illustrated in Figure~\ref{fig:demo_begin}, our method is consistently preferred in human evaluations, confirming its perceptual advantage.

\begin{wrapfigure}{r}{0.4\textwidth}
\vspace{-1.9em}
    \centering 
    \includegraphics[width=0.4\textwidth]{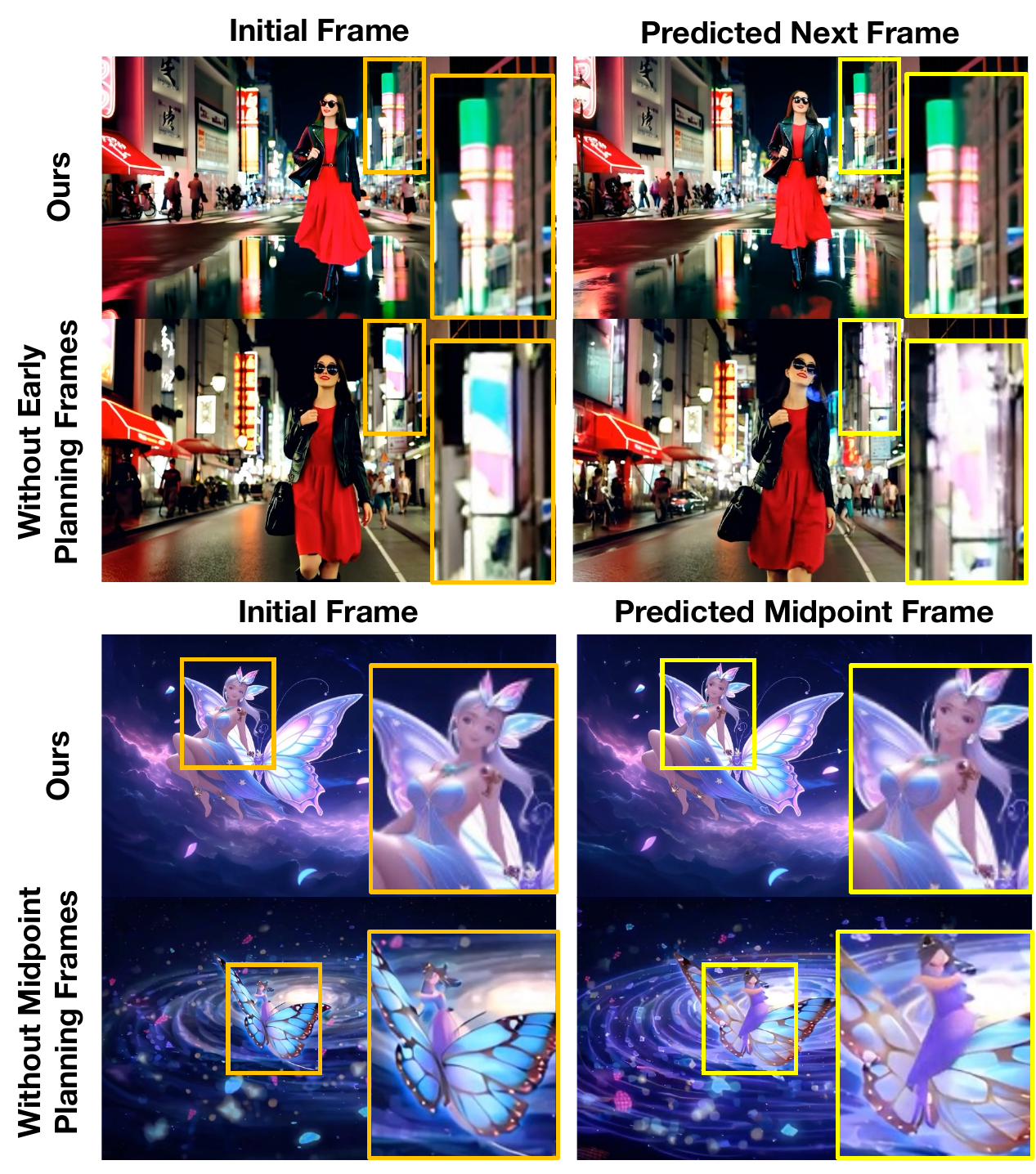}
    \vspace{-0.2in}
    \caption{Qualitative comparisons of different MMPL variants. }
    \label{fig:ab_vis}
\vspace{-2em}
\end{wrapfigure}

\textbf{Qualitative Results.}
As illustrated in Figure~\ref{fig:Qualitative}, AR baselines exhibit severe temporal drift, caused by error accumulation during long-video generation. Over the course of 30-second sequences, these models progressively lose visual fidelity, with artifacts such as blurring, fading, and noticeable color drift becoming increasingly pronounced. The degradation often compounds in dynamic scenes, where motion discontinuities and geometric distortions further undermine temporal coherence. In contrast, our approach sustains high visual quality across the entire sequence, demonstrating strong robustness to motion drift and color distortion. It consistently surpasses CausVid and Self Forcing, and further achieves superior performance to SkyReels-V2 and MAGI-1 under challenging long-horizon conditions, highlighting its effectiveness for stable and high-fidelity long video synthesis.

\textbf{Parallel Inference Efficiency.}
To highlight the practical advantages of Macro-from-Micro Planning, we compare its standard inference with the parallelized variant. 
The parallel strategy achieves substantial speedups without compromising generation quality. 
As illustrated in Figure~\ref{fig:demo_begin}, our method significantly reduces generation time for 60-second videos, demonstrating strong scalability and suitability for real-time deployment. 
Notably, using only two GPUs halves the inference time, and thanks to the pipeline design, four GPUs further reduce the generation time to roughly one-third of the original. 
These results confirm that our approach effectively balances throughput and quality, 
and its hardware efficiency makes it highly amenable to large-scale video synthesis applications.
\begin{table*}[t]
\centering
\scriptsize
\setlength{\tabcolsep}{3pt}
\renewcommand{\arraystretch}{1.05}
\begin{minipage}{0.48\textwidth}
\centering
\caption{Ablation studies on planning setups.}
\label{tab:planning_setups}
\begin{tabular}{l|ccccc}
\toprule
 & \multicolumn{5}{c}{\textit{VBench}} \\
\cmidrule(lr){2-6}
\textbf{Variant} & Subj. & Back. & Mot. & Aes. & Img. \\
\midrule
\rowcolor{gray!10}
\multicolumn{6}{l}{\textit{Planning Setup}} \\
w/o early planning & 0.972 & 0.964 &0.991  & 0.610  & 0.640 \\
w/o midpoint planning &0.977  &0.968  &0.992  &0.618  &0.637  \\
Full             & {0.980} & {0.968} & {0.992} & {0.628} & {0.661} \\
\midrule
\end{tabular}
\end{minipage}
\hfill
\begin{minipage}{0.48\textwidth}
\centering
\caption{Ablation studies on training strategies.}
\label{tab:training_strategy}
\begin{tabular}{l|ccccc}
\toprule
 & \multicolumn{5}{c}{\textit{VBench}} \\
\cmidrule(lr){2-6}
\textbf{Variant} & Subj. & Back. & Mot. & Aes. & Img. \\
\midrule
\rowcolor{gray!10}
\multicolumn{6}{l}{\textit{Training Strategy}} \\
Freeze          & 0.838 & 0.923 & 0.973 & 0.484 & 0.503 \\
Only Q,K     & 0.970 & 0.962 & 0.987  & 0.612 & 0.647 \\
Only Self-Attention              & {0.980} & {0.968} & {0.992} & {0.628} & {0.661} \\
\midrule
\end{tabular}
\end{minipage}
\label{tab:ablation_double}
\end{table*}

\textbf{Ablations on Micro-Planning Frame Placement.}
The placement of Planning frames within each segment during \textit{Micro Planning} is pivotal for MMPL, shaping temporal and structural consistency. We validate this via an ablation with three variants: (i) without early frames (omit frames near the start); (ii) without the midpoint frame (remove the central anchor); and (iii) the full MMPL that retains all Planning frames. As shown in Table~\ref{tab:planning_setups}, the full configuration leads across all metrics. Qualitatively (Figure~\ref{fig:ab_vis}), it yields smoother transitions and more stable long-horizon content, whereas the ablated variants exhibit discontinuities or noticeable jumps around the missing frames.

\textbf{Ablations on Model Training Strategy.}
We compare three update policies for the video generation model: Freeze freezes all parameters; Only Q,K updates only the self-attention query and key projections; Only Self-Attention updates Q, K, V and the attention output, while feed-forward layers and embeddings remain frozen. As shown in Table~\ref{tab:training_strategy}, updating the whole self-attention yields the best scores across all metrics. Training only Q,K is lighter but slightly weaker. Freezing performs worst and shows larger temporal drift and inconsistency. 

\section{Discussion}

\textbf{Compatibility with Acceleration and Distillation Methods.} 
Our paradigm is naturally compatible with acceleration techniques such as DMD and other distillation approaches, requiring no architectural changes. 
During training, we only adjust the attention mask to control the visible frame range, while at inference efficiency is improved by reorganizing the decoding order of video segments. 
This flexibility allows Macro-from-Micro to plug into existing acceleration pipelines. 
The results of adapting model distillation and related strategies are provided in the supplementary material.

\textbf{Compatibility with Self-Forcing Approaches.} 
MMPL also complements self-correcting strategies that mitigate step-wise autoregressive errors, such as Self Forcing. 
In standard training, the model predicts the next frame by denoising conditioned on ground-truth video frames; replacing these with previously generated predictions naturally yields a Self-MMPL regime. 
This hybrid setup extends the duration of generable videos and improves temporal consistency over long sequences.

\textbf{Limitations.}
Although Macro-from-Micro Planning substantially mitigates the accumulation of prediction errors, generating hour-level videos remains challenging due to the reliance on a single static text prompt. The substantial temporal span of long videos means that a single text prompt often aligns only with the early content and fails to capture the full video semantics. Since our current framework does not dynamically update prompts during generation, this static conditioning limits the ability to produce coherent and continuous content across longer segments.

\vspace{-0.5em}
\section{Conclusion}
\vspace{-0.5em}
In this work, we propose a novel planning-then-populating framework centered on Macro-from-Micro Planning (MMPL) for long video generation, which operates without modifying the underlying model architecture. By decomposing video synthesis into Micro Planning, Macro Planning, and Content Populating, MMPL significantly mitigates temporal drift, ensures long-term consistency, and unlocks substantial parallelism in frame generation. Combined with Adaptive Workload Scheduling, it further accelerates long-video generation, reducing inference time to nearly one-third of the original without distillation techniques.
Extensive quantitative and qualitative experiments validate the superior performance of our approach. In the future, we plan to integrate MMPL with model distillation techniques to enable real-time, fully parallelized long video generation, making the framework highly practical for interactive and streaming applications.

\bibliography{2026_conference}

\begin{thebibliography}{45}
\providecommand{\natexlab}[1]{#1}
\providecommand{\url}[1]{\texttt{#1}}
\expandafter\ifx\csname urlstyle\endcsname\relax
  \providecommand{\doi}[1]{doi: #1}\else
  \providecommand{\doi}{doi: \begingroup \urlstyle{rm}\Url}\fi

\bibitem[Arora et~al.(2022)Arora, Asri, Bahuleyan, and Cheung]{Why_Exposure}
Kushal Arora, Layla~El Asri, Hareesh Bahuleyan, and Jackie Chi~Kit Cheung.
\newblock Why exposure bias matters: An imitation learning perspective of error accumulation in language generation.
\newblock In \emph{ACL}, 2022.

\bibitem[Blattmann et~al.(2023)Blattmann, Rombach, Ling, Dockhorn, Kim, Fidler, and Kreis]{alignlatent}
Andreas Blattmann, Robin Rombach, Huan Ling, Tim Dockhorn, Seung~Wook Kim, Sanja Fidler, and Karsten Kreis.
\newblock Align your latents: High-resolution video synthesis with latent diffusion models.
\newblock In \emph{CVPR}, 2023.

\bibitem[Chen et~al.(2024{\natexlab{a}})Chen, Monso, Du, Simchowitz, Tedrake, and Sitzmann]{diffusionforcing}
Boyuan Chen, Diego~Marti Monso, Yilun Du, Max Simchowitz, Russ Tedrake, and Vincent Sitzmann.
\newblock Diffusion forcing: Next-token prediction meets full-sequence diffusion.
\newblock In \emph{NeurIPS}, 2024{\natexlab{a}}.

\bibitem[Chen et~al.(2025)Chen, Lin, Yang, Lin, Zhu, Fan, Zhang, Chen, Chen, Ma, Xiong, Wang, Pang, Kang, Xu, Jin, Liang, Song, Zhao, Xu, Qiu, Li, Fei, Li, and Zhou]{SkyReels}
Guibin Chen, Dixuan Lin, Jiangping Yang, Chunze Lin, Junchen Zhu, Mingyuan Fan, Hao Zhang, Sheng Chen, Zheng Chen, Chengcheng Ma, Weiming Xiong, Wei Wang, Nuo Pang, Kang Kang, Zhiheng Xu, Yuzhe Jin, Yupeng Liang, Yubing Song, Peng Zhao, Boyuan Xu, Di~Qiu, Debang Li, Zhengcong Fei, Yang Li, and Yahui Zhou.
\newblock Skyreels-v2: Infinite-length film generative model.
\newblock \emph{CoRR}, 2025.

\bibitem[Chen et~al.(2023)Chen, Xia, He, Zhang, Cun, Yang, Xing, Liu, Chen, Wang, Weng, and Shan]{videocrafter}
Haoxin Chen, Menghan Xia, Yingqing He, Yong Zhang, Xiaodong Cun, Shaoshu Yang, Jinbo Xing, Yaofang Liu, Qifeng Chen, Xintao Wang, Chao Weng, and Ying Shan.
\newblock Videocrafter1: Open diffusion models for high-quality video generation.
\newblock \emph{CoRR}, 2023.

\bibitem[Chen et~al.(2024{\natexlab{b}})Chen, Xu, Ren, Cong, He, Xie, Sinha, Luo, Xiang, and P{\'{e}}rez{-}R{\'{u}}a]{GenTron}
Shoufa Chen, Mengmeng Xu, Jiawei Ren, Yuren Cong, Sen He, Yanping Xie, Animesh Sinha, Ping Luo, Tao Xiang, and Juan{-}Manuel P{\'{e}}rez{-}R{\'{u}}a.
\newblock Gentron: Diffusion transformers for image and video generation.
\newblock In \emph{CVPR}, 2024{\natexlab{b}}.

\bibitem[Dao et~al.(2022)Dao, Fu, Ermon, Rudra, and R{\'{e}}]{flashattention}
Tri Dao, Daniel~Y. Fu, Stefano Ermon, Atri Rudra, and Christopher R{\'{e}}.
\newblock Flashattention: Fast and memory-efficient exact attention with io-awareness.
\newblock In \emph{NeurIPS}, 2022.

\bibitem[Deng et~al.(2025)Deng, Pan, Diao, Luo, Cui, Lu, Shan, Qi, and Wang]{NOVA}
Haoge Deng, Ting Pan, Haiwen Diao, Zhengxiong Luo, Yufeng Cui, Huchuan Lu, Shiguang Shan, Yonggang Qi, and Xinlong Wang.
\newblock Autoregressive video generation without vector quantization.
\newblock In \emph{ICLR}, 2025.

\bibitem[Dhariwal \& Nichol(2021)Dhariwal and Nichol]{beatsgan}
Prafulla Dhariwal and Alexander~Quinn Nichol.
\newblock Diffusion models beat gans on image synthesis.
\newblock In \emph{NeurIPS}, 2021.

\bibitem[Dong et~al.(2024)Dong, Feng, Guessous, Liang, and He]{flexattention}
Juechu Dong, Boyuan Feng, Driss Guessous, Yanbo Liang, and Horace He.
\newblock Flex attention: {A} programming model for generating optimized attention kernels.
\newblock \emph{CoRR}, 2024.

\bibitem[Gao et~al.(2024)Gao, Shi, Zhang, Wang, and Xiao]{vidgpt}
Kaifeng Gao, Jiaxin Shi, Hanwang Zhang, Chunping Wang, and Jun Xiao.
\newblock Vid-gpt: Introducing gpt-style autoregressive generation in video diffusion models.
\newblock \emph{CoRR}, 2024.

\bibitem[Guo et~al.(2024)Guo, Yang, Rao, Liang, Wang, Qiao, Agrawala, Lin, and Dai]{animatediff}
Yuwei Guo, Ceyuan Yang, Anyi Rao, Zhengyang Liang, Yaohui Wang, Yu~Qiao, Maneesh Agrawala, Dahua Lin, and Bo~Dai.
\newblock Animatediff: Animate your personalized text-to-image diffusion models without specific tuning.
\newblock In \emph{ICLR}, 2024.

\bibitem[Gupta et~al.(2024)Gupta, Yu, Sohn, Gu, Hahn, Li, Essa, Jiang, and Lezama]{WALT}
Agrim Gupta, Lijun Yu, Kihyuk Sohn, Xiuye Gu, Meera Hahn, Fei{-}Fei Li, Irfan Essa, Lu~Jiang, and Jos{\'{e}} Lezama.
\newblock Photorealistic video generation with diffusion models.
\newblock In \emph{ECCV}, 2024.

\bibitem[He et~al.(2024)He, Chen, He, He, Zhou, Zhang, and Zhuang]{zipar}
Yefei He, Feng Chen, Yuanyu He, Shaoxuan He, Hong Zhou, Kaipeng Zhang, and Bohan Zhuang.
\newblock Zipar: Accelerating auto-regressive image generation through spatial locality.
\newblock \emph{CoRR}, 2024.

\bibitem[Ho et~al.(2022)Ho, Salimans, Gritsenko, Chan, Norouzi, and Fleet]{videodiffusionmodels}
Jonathan Ho, Tim Salimans, Alexey~A. Gritsenko, William Chan, Mohammad Norouzi, and David~J. Fleet.
\newblock Video diffusion models.
\newblock In \emph{NeurIPS}, 2022.

\bibitem[Hong et~al.(2023)Hong, Ding, Zheng, Liu, and Tang]{Cogvideo}
Wenyi Hong, Ming Ding, Wendi Zheng, Xinghan Liu, and Jie Tang.
\newblock Cogvideo: Large-scale pretraining for text-to-video generation via transformers.
\newblock In \emph{ICLR}, 2023.

\bibitem[Hu(2024)]{Animateanyone}
Li~Hu.
\newblock Animate anyone: Consistent and controllable image-to-video synthesis for character animation.
\newblock In \emph{CVPR}, 2024.

\bibitem[Huang et~al.(2025{\natexlab{a}})Huang, Li, He, Zhou, and Shechtman]{SelfForcing}
Xun Huang, Zhengqi Li, Guande He, Mingyuan Zhou, and Eli Shechtman.
\newblock Self forcing: Bridging the train-test gap in autoregressive video diffusion.
\newblock \emph{CoRR}, 2025{\natexlab{a}}.

\bibitem[Huang et~al.(2025{\natexlab{b}})Huang, Chen, Ding, Zhang, Dai, Zou, Xiong, and Tian]{IM-Zero}
Yuyang Huang, Yabo Chen, Li~Ding, Xiaopeng Zhang, Wenrui Dai, Junni Zou, Hongkai Xiong, and Qi~Tian.
\newblock Im-zero: Instance-level motion controllable video generation in a zero-shot manner.
\newblock In \emph{CVPR}, 2025{\natexlab{b}}.

\bibitem[Kim et~al.(2024)Kim, Kang, Choi, and Han]{FIFO}
Jihwan Kim, Junoh Kang, Jinyoung Choi, and Bohyung Han.
\newblock Fifo-diffusion: Generating infinite videos from text without training.
\newblock In \emph{NeurIPS}, 2024.

\bibitem[Li et~al.(2025)Li, Hu, Liu, Zhou, Choi, Meng, Guo, Li, Ling, and Wei]{ARLON}
Zongyi Li, Shujie Hu, Shujie Liu, Long Zhou, Jeongsoo Choi, Lingwei Meng, Xun Guo, Jinyu Li, Hefei Ling, and Furu Wei.
\newblock {ARLON:} boosting diffusion transformers with autoregressive models for long video generation.
\newblock In \emph{ICLR}, 2025.

\bibitem[Ma et~al.(2025)Ma, Wang, Chen, Jia, Liu, Li, Chen, and Qiao]{Latte}
Xin Ma, Yaohui Wang, Xinyuan Chen, Gengyun Jia, Ziwei Liu, Yuan{-}Fang Li, Cunjian Chen, and Yu~Qiao.
\newblock Latte: Latent diffusion transformer for video generation.
\newblock \emph{TMLR}, 2025.

\bibitem[Ning et~al.(2024)Ning, Li, Su, Salah, and Ertugrul]{exposure-1}
Mang Ning, Mingxiao Li, Jianlin Su, Albert~Ali Salah, and Itir~{\"{O}}nal Ertugrul.
\newblock Elucidating the exposure bias in diffusion models.
\newblock In \emph{ICLR}, 2024.

\bibitem[Pang et~al.(2025)Pang, Zhang, Luan, Man, Tan, Zhang, Freeman, and Wang]{randar}
Ziqi Pang, Tianyuan Zhang, Fujun Luan, Yunze Man, Hao Tan, Kai Zhang, William~T. Freeman, and Yu{-}Xiong Wang.
\newblock Randar: Decoder-only autoregressive visual generation in random orders.
\newblock In \emph{CVPR}, 2025.

\bibitem[Peebles \& Xie(2023)Peebles and Xie]{DiT}
William Peebles and Saining Xie.
\newblock Scalable diffusion models with transformers.
\newblock In \emph{ICCV}, 2023.

\bibitem[Polyak et~al.(2024)Polyak, Zohar, Brown, Tjandra, Sinha, Lee, Vyas, Shi, Ma, Chuang, Yan, Choudhary, Wang, Sethi, Pang, Ma, Misra, Hou, Wang, Jagadeesh, Li, Zhang, Singh, Williamson, Le, Yu, Singh, Zhang, Vajda, Duval, Girdhar, Sumbaly, Rambhatla, Tsai, Azadi, Datta, Chen, Bell, Ramaswamy, Sheynin, Bhattacharya, Motwani, Xu, Li, Hou, Hsu, Yin, Dai, Taigman, Luo, Liu, Wu, Zhao, Kirstain, He, He, Pumarola, Thabet, Sanakoyeu, Mallya, Guo, Araya, Kerr, Wood, Liu, Peng, Vengertsev, Sch{\"{o}}nfeld, Blanchard, Juefei{-}Xu, Nord, Liang, Hoffman, Kohler, Fire, Sivakumar, Chen, Yu, Gao, Georgopoulos, Moritz, Sampson, Li, Parmeggiani, Fine, Fowler, Petrovic, and Du]{MovieGen}
Adam Polyak, Amit Zohar, Andrew Brown, Andros Tjandra, Animesh Sinha, Ann Lee, Apoorv Vyas, Bowen Shi, Chih{-}Yao Ma, Ching{-}Yao Chuang, David Yan, Dhruv Choudhary, Dingkang Wang, Geet Sethi, Guan Pang, Haoyu Ma, Ishan Misra, Ji~Hou, Jialiang Wang, Kiran Jagadeesh, Kunpeng Li, Luxin Zhang, Mannat Singh, Mary Williamson, Matt Le, Matthew Yu, Mitesh~Kumar Singh, Peizhao Zhang, Peter Vajda, Quentin Duval, Rohit Girdhar, Roshan Sumbaly, Sai~Saketh Rambhatla, Sam~S. Tsai, Samaneh Azadi, Samyak Datta, Sanyuan Chen, Sean Bell, Sharadh Ramaswamy, Shelly Sheynin, Siddharth Bhattacharya, Simran Motwani, Tao Xu, Tianhe Li, Tingbo Hou, Wei{-}Ning Hsu, Xi~Yin, Xiaoliang Dai, Yaniv Taigman, Yaqiao Luo, Yen{-}Cheng Liu, Yi{-}Chiao Wu, Yue Zhao, Yuval Kirstain, Zecheng He, Zijian He, Albert Pumarola, Ali~K. Thabet, Artsiom Sanakoyeu, Arun Mallya, Baishan Guo, Boris Araya, Breena Kerr, Carleigh Wood, Ce~Liu, Cen Peng, Dmitry Vengertsev, Edgar Sch{\"{o}}nfeld, Elliot Blanchard, Felix Juefei{-}Xu, Fraylie Nord, Jeff Liang,
  John Hoffman, Jonas Kohler, Kaolin Fire, Karthik Sivakumar, Lawrence Chen, Licheng Yu, Luya Gao, Markos Georgopoulos, Rashel Moritz, Sara~K. Sampson, Shikai Li, Simone Parmeggiani, Steve Fine, Tara Fowler, Vladan Petrovic, and Yuming Du.
\newblock Movie gen: {A} cast of media foundation models.
\newblock \emph{CoRR}, 2024.

\bibitem[Rombach et~al.(2022)Rombach, Blattmann, Lorenz, Esser, and Ommer]{LDM}
Robin Rombach, Andreas Blattmann, Dominik Lorenz, Patrick Esser, and Bj{\"{o}}rn Ommer.
\newblock High-resolution image synthesis with latent diffusion models.
\newblock In \emph{CVPR}, 2022.

\bibitem[Ross et~al.(2011)Ross, Gordon, and Bagnell]{regret}
St{\'{e}}phane Ross, Geoffrey~J. Gordon, and Drew Bagnell.
\newblock A reduction of imitation learning and structured prediction to no-regret online learning.
\newblock In \emph{AISTATS}, 2011.

\bibitem[Song et~al.(2025)Song, Chen, Simchowitz, Du, Tedrake, and Sitzmann]{historydiffusion}
Kiwhan Song, Boyuan Chen, Max Simchowitz, Yilun Du, Russ Tedrake, and Vincent Sitzmann.
\newblock History-guided video diffusion.
\newblock \emph{CoRR}, 2025.

\bibitem[Sun et~al.(2025)Sun, Wang, Li, Liu, Sun, Feng, Lao, Zhou, He, and Liu]{ardiffusion}
Mingzhen Sun, Weining Wang, Gen Li, Jiawei Liu, Jiahui Sun, Wanquan Feng, Shanshan Lao, SiYu Zhou, Qian He, and Jing Liu.
\newblock Ar-diffusion: Asynchronous video generation with auto-regressive diffusion.
\newblock In \emph{CVPR}, 2025.

\bibitem[Teng et~al.(2025)Teng, Jia, Sun, Li, Li, Tang, Han, Zhang, Zhang, Luo, Kang, Sun, Cao, Huang, Lin, Fang, Tao, Zhang, Wang, Liu, Shi, Su, Sun, Pan, Wang, Sheng, Cui, Hu, Yan, Yin, Zhang, Liu, Yin, Yang, Song, Hu, Zhang, and Li]{MAGI}
Hansi Teng, Hongyu Jia, Lei Sun, Lingzhi Li, Maolin Li, Mingqiu Tang, Shuai Han, Tianning Zhang, W.~Q. Zhang, Weifeng Luo, Xiaoyang Kang, Yuchen Sun, Yue Cao, Yunpeng Huang, Yutong Lin, Yuxin Fang, Zewei Tao, Zheng Zhang, Zhongshu Wang, Zixun Liu, Dai Shi, Guoli Su, Hanwen Sun, Hong Pan, Jie Wang, Jiexin Sheng, Min Cui, Min Hu, Ming Yan, Shucheng Yin, Siran Zhang, Tingting Liu, Xianping Yin, Xiaoyu Yang, Xin Song, Xuan Hu, Yankai Zhang, and Yuqiao Li.
\newblock {MAGI-1:} autoregressive video generation at scale.
\newblock \emph{CoRR}, 2025.

\bibitem[Wang et~al.(2025{\natexlab{a}})Wang, Ai, Wen, Mao, Xie, Chen, Yu, Zhao, Yang, Zeng, Wang, Zhang, Zhou, Wang, Chen, Zhu, Zhao, Yan, Huang, Meng, Zhang, Li, Wu, Chu, Feng, Zhang, Sun, Fang, Wang, Gui, Weng, Shen, Lin, Wang, Wang, Zhou, Wang, Shen, Yu, Shi, Huang, Xu, Kou, Lv, Li, Liu, Wang, Zhang, Huang, Li, Wu, Liu, Pan, Zheng, Hong, Shi, Feng, Jiang, Han, Wu, and Liu]{Wan}
Ang Wang, Baole Ai, Bin Wen, Chaojie Mao, Chen{-}Wei Xie, Di~Chen, Feiwu Yu, Haiming Zhao, Jianxiao Yang, Jianyuan Zeng, Jiayu Wang, Jingfeng Zhang, Jingren Zhou, Jinkai Wang, Jixuan Chen, Kai Zhu, Kang Zhao, Keyu Yan, Lianghua Huang, Xiaofeng Meng, Ningyi Zhang, Pandeng Li, Pingyu Wu, Ruihang Chu, Ruili Feng, Shiwei Zhang, Siyang Sun, Tao Fang, Tianxing Wang, Tianyi Gui, Tingyu Weng, Tong Shen, Wei Lin, Wei Wang, Wei Wang, Wenmeng Zhou, Wente Wang, Wenting Shen, Wenyuan Yu, Xianzhong Shi, Xiaoming Huang, Xin Xu, Yan Kou, Yangyu Lv, Yifei Li, Yijing Liu, Yiming Wang, Yingya Zhang, Yitong Huang, Yong Li, You Wu, Yu~Liu, Yulin Pan, Yun Zheng, Yuntao Hong, Yupeng Shi, Yutong Feng, Zeyinzi Jiang, Zhen Han, Zhi{-}Fan Wu, and Ziyu Liu.
\newblock Wan: Open and advanced large-scale video generative models.
\newblock \emph{CoRR}, 2025{\natexlab{a}}.

\bibitem[Wang et~al.(2025{\natexlab{b}})Wang, Ren, Lin, Han, Guo, Yang, Zou, Feng, and Liu]{PAR}
Yuqing Wang, Shuhuai Ren, Zhijie Lin, Yujin Han, Haoyuan Guo, Zhenheng Yang, Difan Zou, Jiashi Feng, and Xihui Liu.
\newblock Parallelized autoregressive visual generation.
\newblock In \emph{CVPR}, 2025{\natexlab{b}}.

\bibitem[Wu et~al.(2025{\natexlab{a}})Wu, Gao, Poole, Trevithick, Zheng, Barron, and Holynski]{CAT4D}
Rundi Wu, Ruiqi Gao, Ben Poole, Alex Trevithick, Changxi Zheng, Jonathan~T. Barron, and Aleksander Holynski.
\newblock {CAT4D:} create anything in 4d with multi-view video diffusion models.
\newblock In \emph{CVPR}, 2025{\natexlab{a}}.

\bibitem[Wu et~al.(2025{\natexlab{b}})Wu, Yang, Po, Xu, Liu, Lin, and Wetzstein]{WorldV}
Tong Wu, Shuai Yang, Ryan Po, Yinghao Xu, Ziwei Liu, Dahua Lin, and Gordon Wetzstein.
\newblock Video world models with long-term spatial memory.
\newblock \emph{CoRR}, 2025{\natexlab{b}}.

\bibitem[Xiang et~al.(2025)Xiang, Xue, Dai, Wang, Li, Yue, Ma, Yu, Chang, and Yu]{ReMask}
Xunzhi Xiang, Haiwei Xue, Zonghong Dai, Di~Wang, Minglei Li, Ye~Yue, Fei Ma, Weijiang Yu, Heng Chang, and Fei~Richard Yu.
\newblock Remask-animate: Refined character image animation using mask-guided adapters.
\newblock In \emph{AAAI}, 2025.

\bibitem[Yan et~al.(2021)Yan, Zhang, Abbeel, and Srinivas]{VideoGPT}
Wilson Yan, Yunzhi Zhang, Pieter Abbeel, and Aravind Srinivas.
\newblock Videogpt: Video generation using {VQ-VAE} and transformers.
\newblock \emph{CoRR}, 2021.

\bibitem[Yin et~al.(2025)Yin, Zhang, Zhang, Freeman, Durand, Shechtman, and Huang]{Causvid}
Tianwei Yin, Qiang Zhang, Richard Zhang, William~T. Freeman, Fr{\'{e}}do Durand, Eli Shechtman, and Xun Huang.
\newblock From slow bidirectional to fast causal video generators.
\newblock In \emph{CVPR}, 2025.

\bibitem[Yu et~al.(2024)Yu, Lezama, Gundavarapu, Versari, Sohn, Minnen, Cheng, Gupta, Gu, Hauptmann, Gong, Yang, Essa, Ross, and Jiang]{LBD}
Lijun Yu, Jos{\'{e}} Lezama, Nitesh~Bharadwaj Gundavarapu, Luca Versari, Kihyuk Sohn, David Minnen, Yong Cheng, Agrim Gupta, Xiuye Gu, Alexander~G. Hauptmann, Boqing Gong, Ming{-}Hsuan Yang, Irfan Essa, David~A. Ross, and Lu~Jiang.
\newblock Language model beats diffusion - tokenizer is key to visual generation.
\newblock In \emph{ICLR}, 2024.

\bibitem[Zhang et~al.(2024)Zhang, Yuanzhi, Xi, Fangqiu, and Xuelong]{zhang2024vast}
Chi Zhang, LiangLi Yuanzhi, Qiu Xi, Yi~Fangqiu, and Li~Xuelong.
\newblock Vast 1.0: A unified framework for controllable and consistent video generation.
\newblock \emph{CoRR}, 2024.

\bibitem[Zhang et~al.(2025)Zhang, Shi, Jiang, Xiang, Qian, Shi, and Jiang]{Proteus}
Guiyu Zhang, Chen Shi, Zijian Jiang, Xunzhi Xiang, Jingjing Qian, Shaoshuai Shi, and Li~Jiang.
\newblock Proteus-id: Id-consistent and motion-coherent video customization.
\newblock \emph{CoRR}, 2025.

\bibitem[Zhang \& Agrawala(2025)Zhang and Agrawala]{zhang2025framepack}
Lvmin Zhang and Maneesh Agrawala.
\newblock Packing input frame contexts in next-frame prediction models for video generation.
\newblock \emph{CoRR}, 2025.

\bibitem[Zhao et~al.(2025)Zhao, Liu, Wang, Chen, Wang, Chen, Zhang, and Shen]{MovieDreamer}
Canyu Zhao, Mingyu Liu, Wen Wang, Weihua Chen, Fan Wang, Hao Chen, Bo~Zhang, and Chunhua Shen.
\newblock Moviedreamer: Hierarchical generation for coherent long visual sequences.
\newblock In \emph{ICLR}, 2025.

\bibitem[Zheng et~al.(2025)Zheng, Huang, Liu, Zou, He, Zhang, Zhang, He, Zheng, Qiao, and Liu]{vbench2}
Dian Zheng, Ziqi Huang, Hongbo Liu, Kai Zou, Yinan He, Fan Zhang, Yuanhan Zhang, Jingwen He, Wei{-}Shi Zheng, Yu~Qiao, and Ziwei Liu.
\newblock Vbench-2.0: Advancing video generation benchmark suite for intrinsic faithfulness.
\newblock \emph{CoRR}, 2025.

\bibitem[Zhu et~al.(2024)Zhu, Chen, Dai, Dong, Xu, Cao, Yao, Zhu, and Zhu]{champ}
Shenhao Zhu, Junming~Leo Chen, Zuozhuo Dai, Zilong Dong, Yinghui Xu, Xun Cao, Yao Yao, Hao Zhu, and Siyu Zhu.
\newblock Champ: Controllable and consistent human image animation with 3d parametric guidance.
\newblock In \emph{ECCV}, 2024.

\end{thebibliography}
\bibliographystyle{2026_conference}

\newpage
\appendix

\section{Data Preparation}
\textbf{Data Sources and Filtering.}
Our dataset comprises two components: (i) licensed commercial videos purchased from authorized providers; and (ii) web videos manually collected from open platforms, primarily \href{https://mixkit.co/free-stock-video/}{Mixkit}, \href{https://www.pexels.com/zh-cn/}{Pexels}, and \href{https://pixabay.com/videos/}{Pixabay}. 
For all candidate videos, we first generate textual captions with Qwen-72B and compute an aesthetic score using the LAION aesthetic predictor. 
For licensed data, we rank by the aesthetic score and retain only the top 1\%. 
For web-collected data, we conduct human quality control to remove low-quality clips, content--caption mismatches, and potential copyright risks, and to verify caption consistency. 
We then merge the two subsets, yielding approximately $\sim$50k high-quality samples to train our long-video generation model.

\textbf{Structured Annotation Pipeline.}
To obtain rich and structured annotations, we drive Qwen-72B with carefully designed instruction prompts to analyze video frames and output a JSON object with a fixed schema. 
The JSON includes: a short scene summary (\texttt{short\_caption}); a dense contextual description (\texttt{dense\_caption}, covering main subject, background, visual style, camera movement, shot type, lighting, and atmosphere); detailed subject descriptions (for persons: facial expressions, emotional state, and ethnicity); background information; standardized style/shot/movement labels; aesthetic tags; and role statistics (e.g., number of humanoid characters, coverage extent, depiction style, and motion dynamics). 
Concretely, \texttt{short\_caption} is generated with the instruction \textit{"Brief scene summary in 1 sentence"}, while \texttt{dense\_caption} uses \textit{"Detailed context including main subject, background, visual style, camera movement tech, shot type, lighting, and atmosphere"}. 
All outputs are in English, follow predefined field orders and constraints, and employ standardized vocabulary for key attributes.

\begin{figure*}[h]
    \centering
    \includegraphics[width=\textwidth]{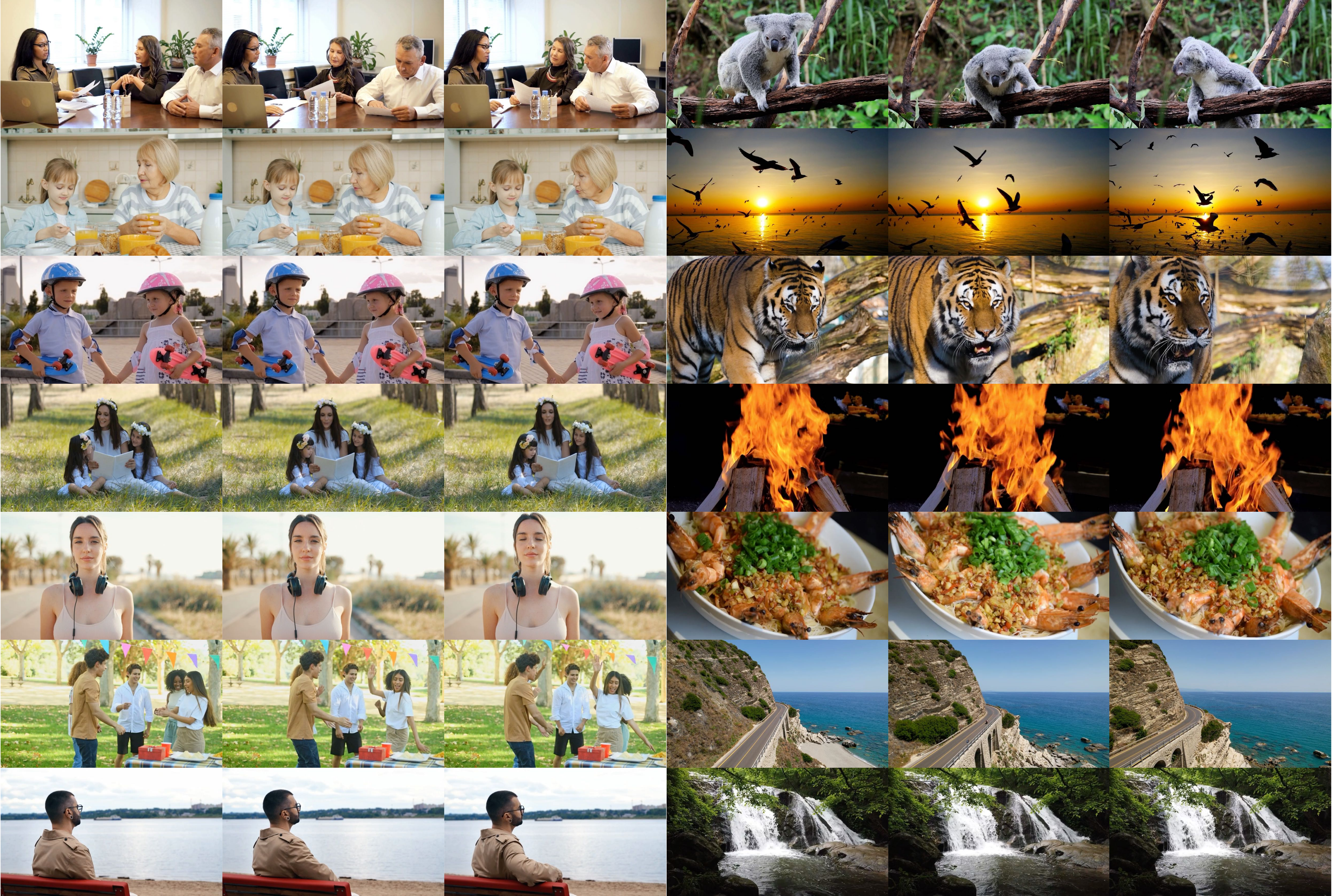}
    \vspace{-0.2in}
    \caption{\textbf{Examples of training samples. }
    The dataset combines licensed and web-collected videos, curated via aesthetic scoring and manual screening.}
    \label{fig:examplesdata}
    \vspace{-0.1in}
\end{figure*}

\textbf{Quality Control and Training Setup.}
After annotation, we re-evaluate aesthetic quality with the LAION predictor to ensure consistency. 
During training, for each of the $\sim$50k videos we sample the conditioning text with probability 0.8 from \texttt{dense\_caption} and with probability 0.2 from \texttt{short\_caption}. 
This strategy preserves the high information density of dense captions while maintaining robustness and diversity from concise summaries. As shown in Figure~\ref{fig:examplesdata}, we present several representative examples from our curated dataset.

\newpage
\section{User Study Details}
\vspace{-1em}
To complement the quantitative metrics, we conduct a user study on long-video generation. In each trial, participants evaluate five videos generated from the same prompt by ranking them (1 = best, 5 = worst) along three dimensions: \emph{text–visual alignment}, \emph{content consistency}, and \emph{long-sequence color stability}. In addition, participants select a single overall favorite video. 
\begin{figure*}[h]
    \centering
    \includegraphics[width=0.9\textwidth]{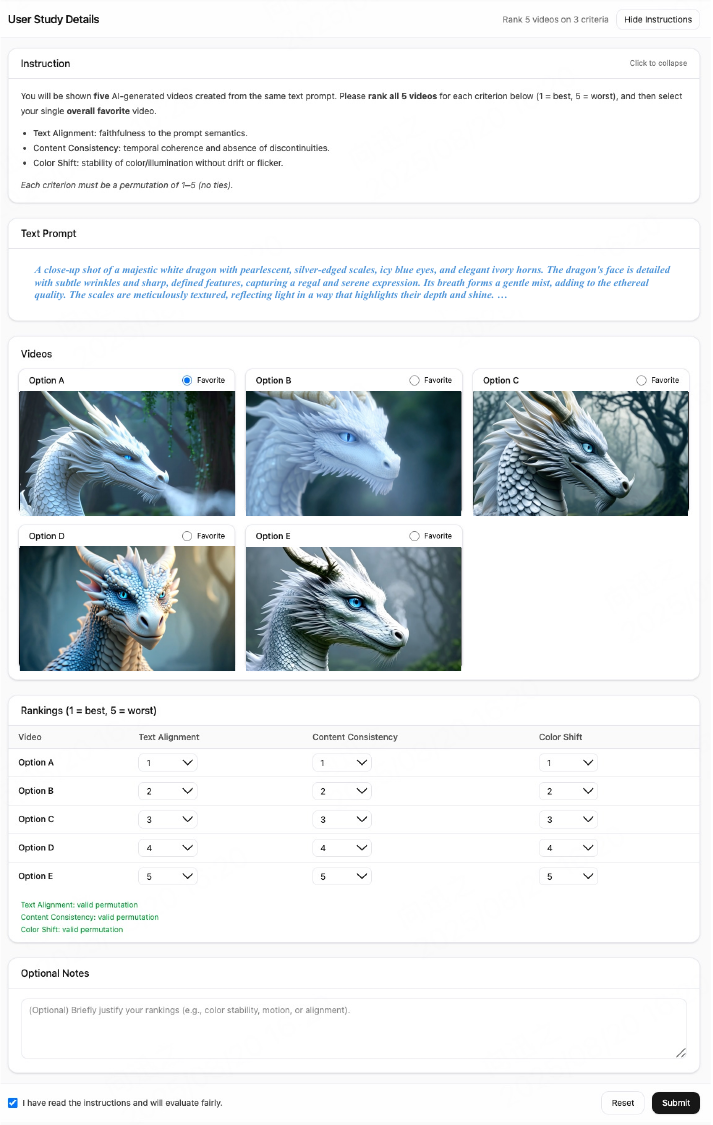}
    \caption{\textbf{User study instruction screenshots.}}
    \label{fig:us}
    \vspace{-0.1in}
\end{figure*}

This protocol provides fine-grained human judgments on both quality and temporal robustness that are not fully captured by automated metrics. Detailed instructions are shown in Figure \ref{fig:us}.

\section{Training Settings}

\subsection{Hyperparameter Settings}
Most experiments are conducted on 32 NVIDIA GPUs (80 GB each), using a per-GPU batch size of 1 without gradient accumulation. The detailed hyperparameters are summarized in Table~\ref{tab:training-hyperparameters}. Training the Teacher Forcing 14B model for 8,000 steps required about three days, while the DMD 1.3B model reached 8,000 steps within roughly one day.

\begin{table}[h]
\scriptsize
\caption{Specification of training hyperparameters}
\label{tab:training-hyperparameters}
\centering
\renewcommand{\arraystretch}{1.2}
\resizebox{\textwidth}{!}{
\begin{tabularx}{\textwidth}{l >{\centering\arraybackslash}X >{\centering\arraybackslash}X}
\toprule
\textbf{Hyperparameters} & \textbf{Teacher Forcing} & \textbf{Self Forcing} \\
\midrule
Generate network & Wan2.1-T2V-14B & Wan2.1-T2V-1.3B \\
Real score network & N/A & Wan2.1-T2V-14B \\
Fake score network & N/A & Wan2.1-T2V-14B \\
Batch size & 32 & 32 \\
Optimizer ($G_\theta$) &
\makecell[c]{AdamW, $\beta_1=0$, $\beta_2=0.999$,\\
$\epsilon=1\!\times\!10^{-8}$, weight\_decay $=0.01$} &
\makecell[c]{Adam, $\beta_1=0$, $\beta_2=0.999$,\\
$\epsilon=1\!\times\!10^{-8}$, weight\_decay $=0.01$} \\
Optimizer ($f_\psi$) & N/A &
\makecell[c]{Adam, $\beta_1=0$, $\beta_2=0.999$,\\
$\epsilon=1\!\times\!10^{-8}$, weight\_decay $=0.01$} \\
Learning rate ($G_\theta$) & $1\!\times\!10^{-5}$ & $2\!\times\!10^{-6}$ \\
Learning rate ($f_\psi$) & N/A & $4\!\times\!10^{-7}$ \\
Gen./Cri. update ratio & N/A & 5 \\
EMA decay & N/A & 0.99 \\
\bottomrule
\end{tabularx}
}
\end{table}

\subsection{Planning Settings}

\begin{figure*}[h]
    \centering
    \includegraphics[width=\textwidth]{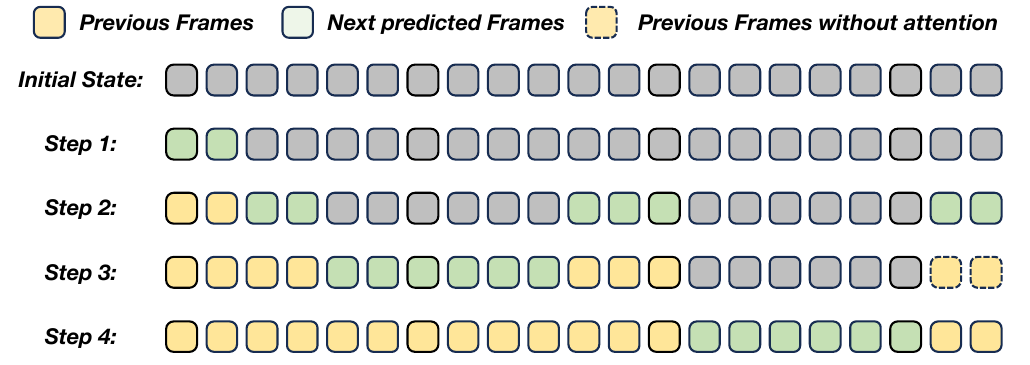}
    \vspace{-0.05in}
    \caption{\textbf{Overview of our planning-based inference on an 81-frame sequence with 21 latent tokens (full Wan model)}. Pre-Planning latent tokens at the beginning,  midpoint, and terminal positions serve as stable anchors in the denoising schedule, guiding the synthesis of all intermediate frames and ensuring long-range temporal coherence.}
    \label{fig:full_me}
    \vspace{-0.1in}
\end{figure*}

\textbf{Settings.}
To clarify the generation process, we detail the model’s computation using the 21 latent tokens of the full Wan model as shown in Figure~\ref{fig:full_me}. Tokens are indexed from 0: indices 0–1 correspond to the initial frame, indices 2–3 to early planning frames, indices 10–12 to midpoint planning frames, and indices 19–20 to terminal planning frames.

\textbf{Analysis.}
While the proposed planning setup places anchors over long horizons, a central challenge remains: enabling the autoregressive (AR) decoder to effectively exploit these anchors when synthesizing intermediate frames. Relying on a single planning token at the \emph{early}, \emph{midpoint}, and \emph{terminal} boundaries is intrinsically fragile in the presence of the AR decoder’s pronounced \emph{recency bias}---the tendency to overweight the most recent observations while underutilizing distant context. This bias causes the model, at each sub-segment junction, to condition predominantly on the tail of the preceding sub-segment, thereby inheriting and amplifying residual errors and inducing cross-boundary propagation. Consequently, a single planning token per boundary is insufficient to arrest drift arising from accumulated errors. Formally, this bias in MMPL is expressed in Eq.~\ref{recency bias}:
\begin{equation}
\label{recency bias}
p\big(x^{s_k+1:e_k-1} \mid {x}^{1:s_k}, x^{e_k}\big)
\approx
p\big(x^{s_k+1:e_k-1} \mid x^{s_{k}-K:s_k}, x^{e_k}\big).
\end{equation}
Here, $s_k$ and $e_k$ denote the starting and ending reference indices of sub-segment $k$, corresponding to the pre-planned \emph{planning frames}. The hyperparameter $K$ specifies the size of the recent-context window on which the AR decoder conditions---namely, the $K$ frames immediately preceding $s_k$. Because $\{s_{k}-K:s_k\}$ overlaps with the tail of the previous sub-segment, residual errors inevitably leak into the current one, leading to error propagation across boundaries.

To counteract this bias, we replace the single predecessor at each boundary with a \emph{local multi-frame set}. Concretely,
\[
\mathcal{P}_{s_1} = \{2,3\}, \quad
\mathcal{P}_{e_1} = \{10,11,12\}, \quad
\mathcal{P}_{s_2} \approx \{10,11,12\}, \quad
\mathcal{P}_{e_2} = \{19,20\},
\]
where $\mathcal{P}_{s_k}$ denotes the local index set of expanded pre-planning frames around boundary $s_k$. Using these expanded anchors, the conditional distribution for sub-segment $k$ is refined to
\begin{equation}
p\big(x^{s_k+1:e_k-1} \mid {x}^{1:s_k}, x^{e_k}\big)
\approx
p\big(x^{s_k+1:e_k-1} \mid x^{(s_{k}-2-K):(s_k-3)}, \, x^{s_{k}-2:s_k}, \, x^{e_k}\big).
\end{equation}
Conditioning on a compact bundle of early-step, low-drift frames—rather than a single predecessor—dilutes residual errors inherited from the previous sub-segment. At the same time, the model’s recency bias naturally prioritizes the most recent elements within this bundle, thereby stabilizing long-horizon synthesis and suppressing cross-boundary error propagation without discarding information from the planned anchors.

\section{Error Accumulation Analysis in AR models}
\label{Analysis}

\textbf{Autoregressive (AR) Models.}
Autoregressive (AR) models generate a sequence $x=(x^1,\dots,x^T)$ by factorizing its joint probability distribution according to the chain rule of probability:
\begin{equation}
p_\theta(x)=\prod_{t=1}^{T}p_\theta(x^t\mid x^{<t}),
\end{equation}
where $x^{<t}=(x^1,\dots,x^{t-1})$ denotes all previously generated elements.
In practice, AR models are commonly trained with the \textit{teacher forcing} strategy, which replaces the model’s own past predictions with the ground-truth history during training.
This reduces the training objective to a standard negative log-likelihood (NLL) minimization:
\begin{equation}
\mathcal{L}(\theta)=-\sum_{t=1}^{T}\log p_\theta(x^t\mid x^{<t}{\text{gt}}),
\end{equation}
where $x^{<t}{\text{gt}}$ denotes the ground-truth prefix of the sequence.
Such training ensures stable and efficient optimization, but it also introduces a train-test discrepancy—commonly referred to as \textit{exposure bias} \citep{exposure-1}—because the model will rely on its own predictions rather than ground-truth history during inference, potentially leading to error accumulation over long sequences.  

To analyze the underlying sources and impacts of error accumulation, we follow \citep{Why_Exposure} and formulate AR generation as a sequential decision process under the imitation learning (IL) framework. 
Here, the state is defined as $s^t=x^{<t}$, the action as $a^t=x^t$, the policy as $\pi_\theta(a^t\mid s^t)=p_\theta(x^t\mid x^{<t})$, and the oracle policy as $\pi^*(a^t\mid s^t)=p_\text{data}(x^t\mid x^{<t})$.
Maximum-likelihood training corresponds to behavior cloning, which minimizes training loss on the oracle-induced state distribution but suffers from compounding errors once the policy is executed on its own rollouts.

In the imitation learning literature \citep{regret}, rolling out a policy trained via behavior cloning often leads to error accumulation. This happens because the policy is executed on its own predictions rather than the oracle states seen during training. To analyze this effect, researchers use inference-time regret, which measures the performance gap between the behavior cloning policy $\pi_{BC}$ and the oracle policy $o$ during rollout:
\begin{equation}
\mathcal{R}(\pi_{BC})=L^I(\pi_{BC})-L^I(o).
\end{equation}

Here, $L^I(\pi)$ denotes the expected cumulative loss (or cost) when executing policy $\pi$ over the entire rollout horizon. 
Let $\epsilon$ denote the average expected error of executing the behavior cloning policy $\pi_{BC}$ over $T$ steps, which itself is upper-bounded. The regret of behavior cloning is bounded by
\begin{equation}
T\epsilon \le \mathcal{R}(\pi_{BC}) \le T^2\epsilon,
\end{equation}
Building on this analysis, and following \citep{Why_Exposure}, we further extend it to the AR video generation setting with model $p_\theta$ and decoding strategy $\mathcal{F}$, which yields
\begin{equation}
T\epsilon \le \mathcal{R}(p_\theta,\mathcal{F}) \le T^2\epsilon,
\end{equation}
which demonstrates that even small per-step errors can accumulate linearly in expectation and quadratically in the worst case, thereby explaining the progressive drift and long-horizon degradation observed in autoregressive generation under exposure bias.

\section{Importance of VAE}
\begin{figure*}[h]
    \centering
    \includegraphics[width=\textwidth]{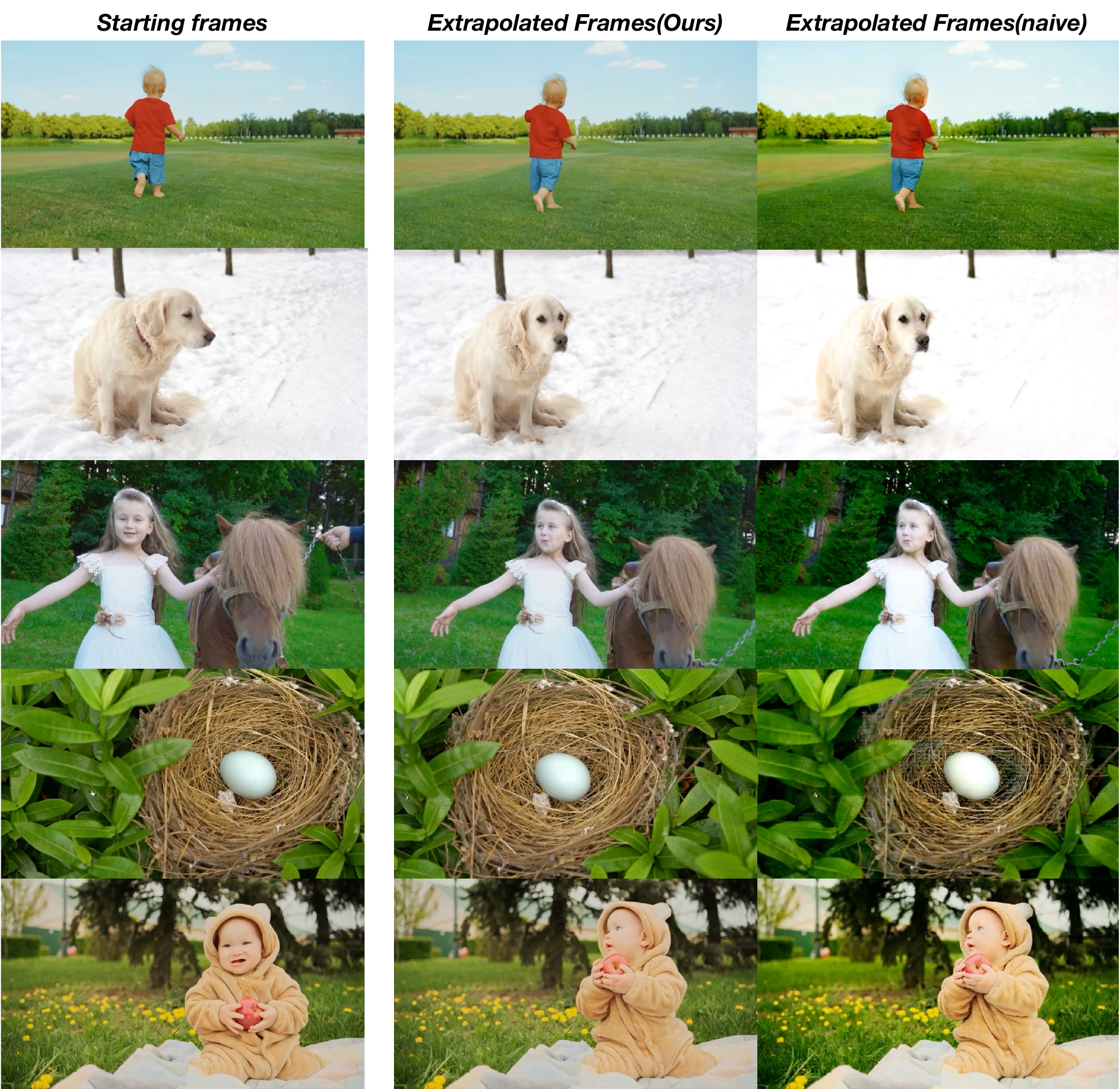}
    \vspace{-0.2in}
    \caption{\textbf{Qualitative comparisons on video extrapolation.}}
    \label{fig:extrapolation}
    \vspace{-0.1in}
\end{figure*}

We compare the extrapolation procedure from the public Causvid \citep{Causvid} codebase against our proposed \emph{drift-resilient re-encoding and decoding strategy} as shown in Figure~\ref{fig:extrapolation}. When extrapolation goes beyond the training context length and requires segment stitching, the baseline suffers from severe color drift and visual artifacts, whereas our method effectively mitigates these degradations.

\section{Noise Initialization Strategy for Smoothing Generation}
\begin{figure*}[h]
    \centering
    \includegraphics[width=\textwidth]{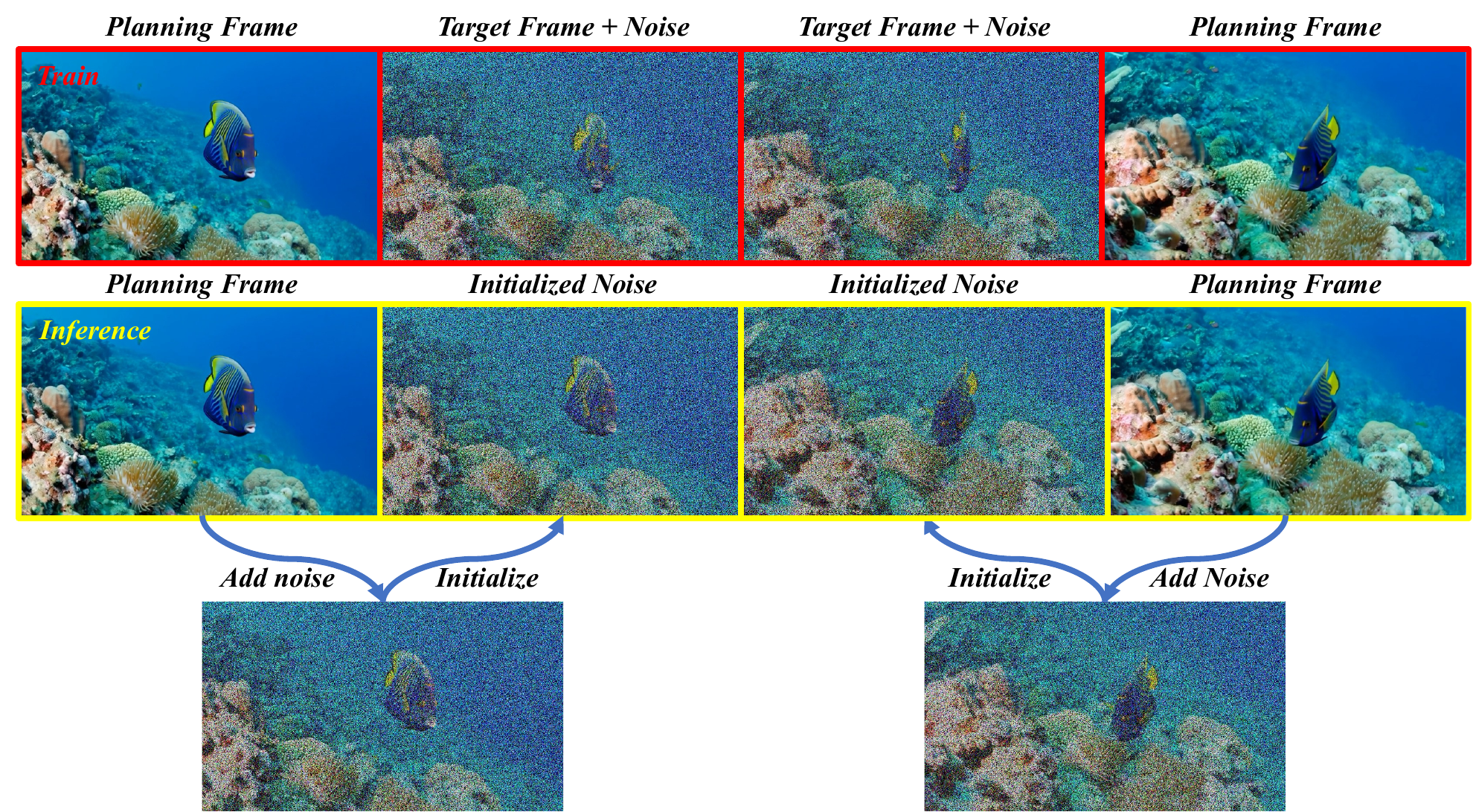}
    \vspace{-0.2in}
    \caption{Framework for stable and smoothing frame generation via coherent noise initialization.}
    \label{fig:Initialized}
    \vspace{-0.1in}
\end{figure*}

In this work, we propose a specialized noise initialization strategy to address potential temporal discontinuities and instability at the transition boundaries between planning frames and content frames as shown in Figure~\ref{fig:Initialized}. This approach ensures smooth visual transitions by strategically incorporating noise information from adjacent planning frames during the content frame generation process. Let $P_{n-1}$ and $P_n$ represent the planning frames at temporal positions $n-1$ and $n$, respectively, and let $C_{n+1}$ denote the target content frame at position $n+1$. To establish the theoretical foundation, we first recall the standard diffusion forward process formulation. Given a clean frame $\mathbf{x}_0$ at diffusion timestep $t$, the noisy observation $\mathbf{x}_t$ is generated through the Gaussian perturbation:
\begin{equation}
q(\mathbf{x}_{t} | \mathbf{x}_{0}) = \mathcal{N}(\mathbf{x}_{t}; \sqrt{\bar{\alpha}{t}} \mathbf{x}_{0}, (1 - \bar{\alpha}{t})\mathbf{I}),
\end{equation}
where $\bar{\alpha}_t$ denotes the cumulative product of the noise schedule coefficients. This process can be equivalently expressed as:
\begin{equation}
\mathbf{x}_{t} = \sqrt{\bar{\alpha}{t}} \cdot \mathbf{x}_{0} + \sqrt{1 - \bar{\alpha}{t}} \cdot \epsilon, \quad \epsilon \sim \mathcal{N}(\mathbf{0}, \mathbf{I}).
\end{equation}
Building upon this formulation, our methodology initializes the noise vector $\epsilon_{C_{n+1}}$ for the content frame $C_{n+1}$ through a weighted interpolation of the noise vectors associated with the preceding planning frames. Specifically, the initialization follows:
\begin{equation}
\epsilon_{C_{n+1}} = \alpha \cdot \epsilon_{P_{n}} + (1 - \alpha) \cdot \epsilon_{P_{n-1}},
\end{equation}
where $\epsilon_{C_{n+1}}$ represents the noise vector utilized in the reverse diffusion process for generating content frame $C_{n+1}$, $\epsilon_{P_n}$ and $\epsilon_{P_{n-1}}$ correspond to the noise vectors derived from planning frames $P_n$ and $P_{n-1}$
This noise initialization strategy ensures a continuous evolution of stochastic patterns across frame boundaries, effectively mitigating visual artifacts and temporal inconsistencies. By controlling the interpolation weight $\alpha$, our method provides precise adjustment over the temporal smoothness characteristics, enabling stable and coherent video generation while maintaining high visual quality throughout the sequence.

\vspace{-0.1in}

\newpage
\section{Scalability}

\subsection{Image-to-Video Extension}
\begin{figure*}[h]
    \centering
    \includegraphics[width=\textwidth]{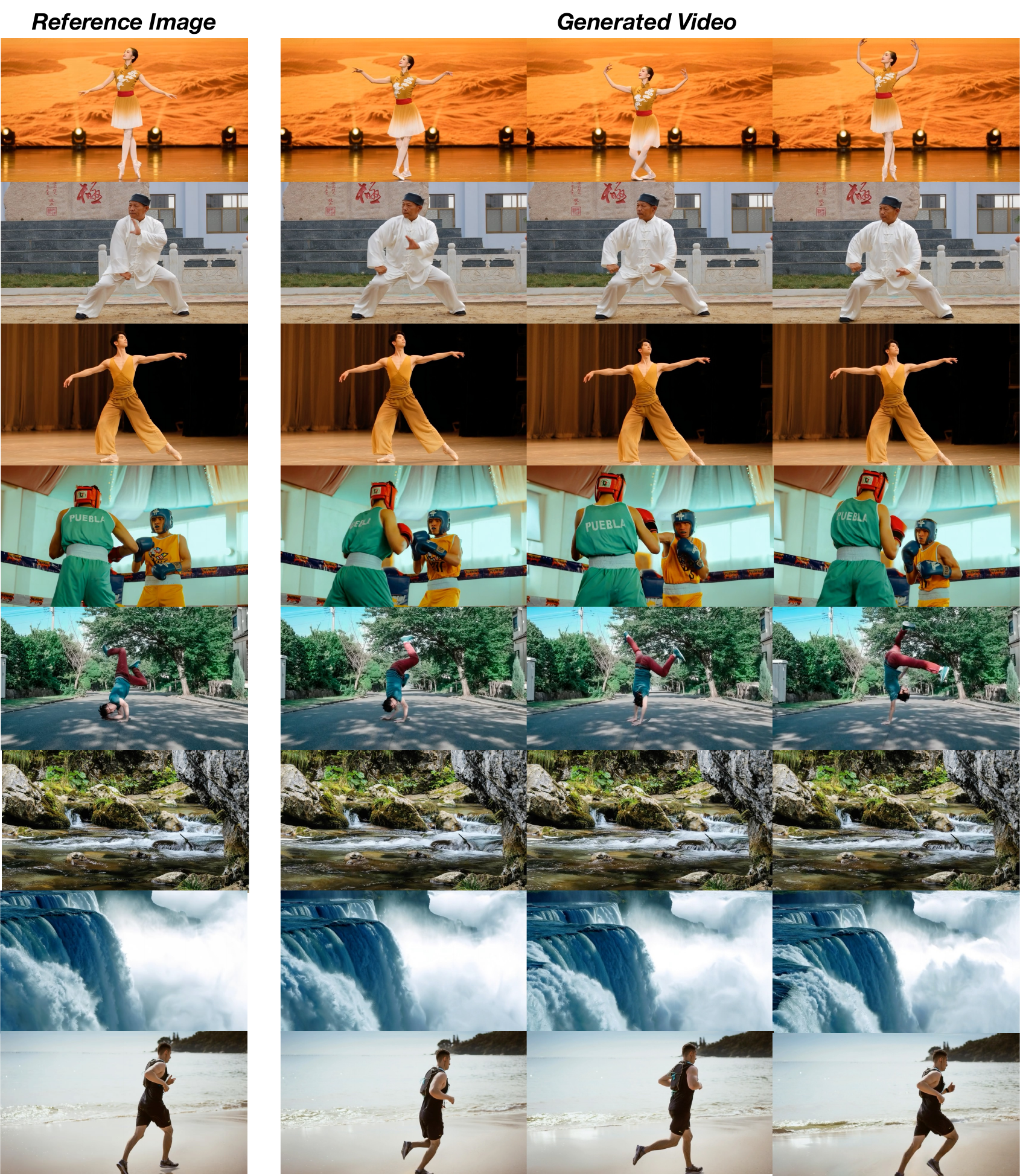}
    \vspace{-0.2in}
    \caption{\textbf{Qualitative results of extending our unified autoregressive framework from text-to-video (T2V) to image-to-video (I2V) generation.} Without any architectural modifications or additional image encoders, the framework adapts seamlessly by only adjusting the number and ordering of autoregressive steps, while preserving temporal consistency and visual quality.}
    \label{fig:QualitativeI2V}
    \vspace{-0.1in}
\end{figure*}

Our framework is not restricted to the text-to-video (T2V) task; it can be seamlessly extended to image-to-video (I2V) generation without introducing any architectural modifications or additional image encoders. This flexibility derives from the unified autoregressive design, which only requires lightweight adjustments to the number and ordering of autoregressive steps. As a result, the framework adapts naturally to different input modalities while maintaining temporal consistency and generation quality as shown in Figure~\ref{fig:QualitativeI2V}.

\subsection{Adaptation to Self-Forcing and DMD}
\begin{figure*}[h]
    \centering
    \includegraphics[width=\textwidth]{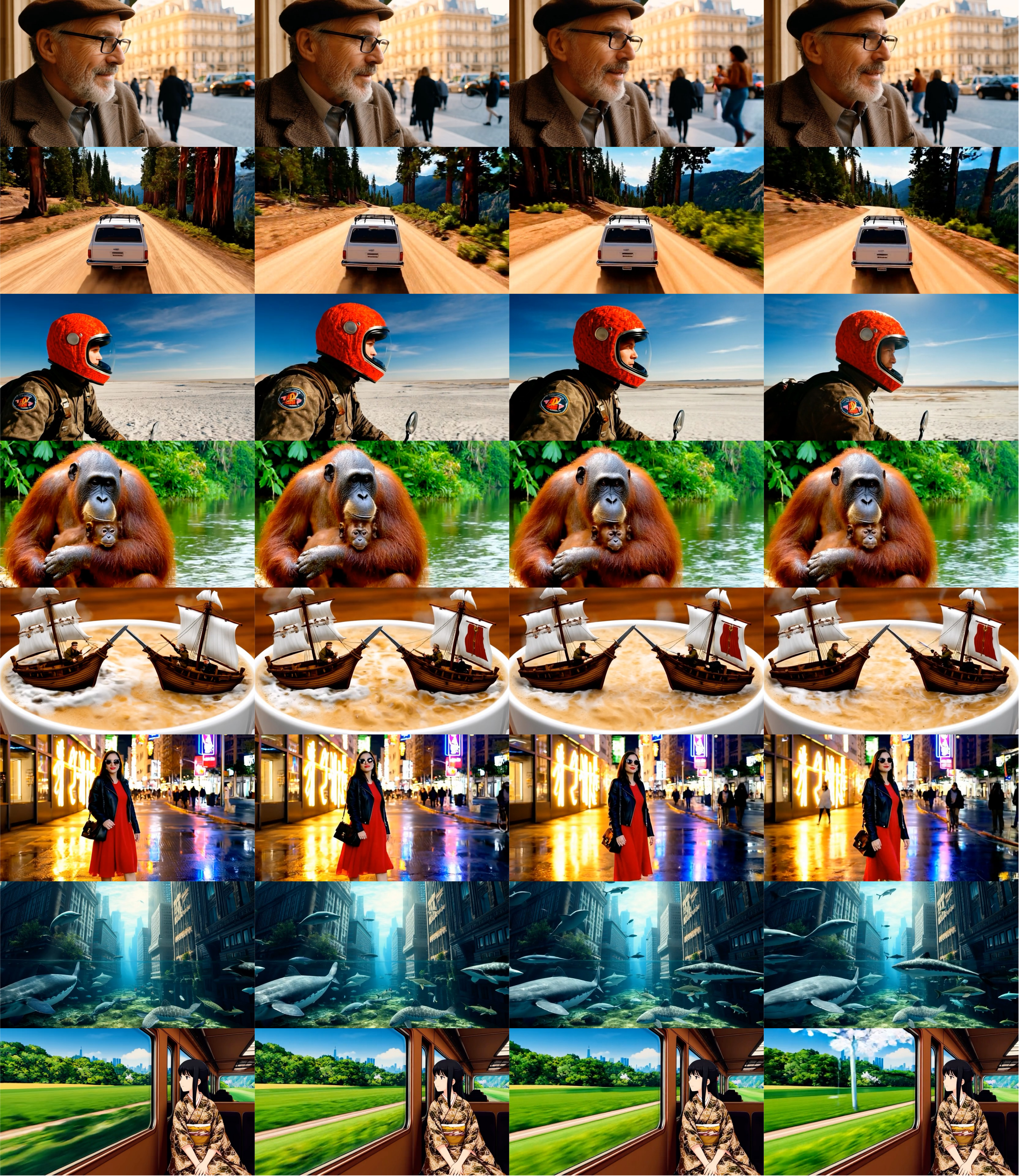}
    \vspace{-0.2in}
    \caption{\textbf{Integration of our framework with Self Forcing \citep{SelfForcing} and DMD \citep{Causvid} strategies}. The adaptation requires no architectural changes—only modifications to the attention visibility range and prediction order during training and inference. Combined with parallelized decoding, the method achieves substantial inference acceleration, sustaining over 32 FPS in long-horizon video generation.}
    \label{fig:QualitativeDMD}
    \vspace{-0.1in}
\end{figure*}

Our approach can be seamlessly integrated with self-forcing strategies without any architectural modifications. 
Specifically, it only requires adjusting the attention visibility range and the prediction order during both training and inference. 
This lightweight adaptation enables direct compatibility with existing self-forcing pipelines, while retaining the benefits of our planning-based design. 
Combined with parallelized decoding, the resulting system achieves substantial inference speedups, sustaining over 32 FPS in long-horizon video generation as shown in Figure~\ref{fig:QualitativeDMD}.

\section{More Qualitative Results}
\begin{figure*}[h]
    \centering
    \includegraphics[width=\textwidth]{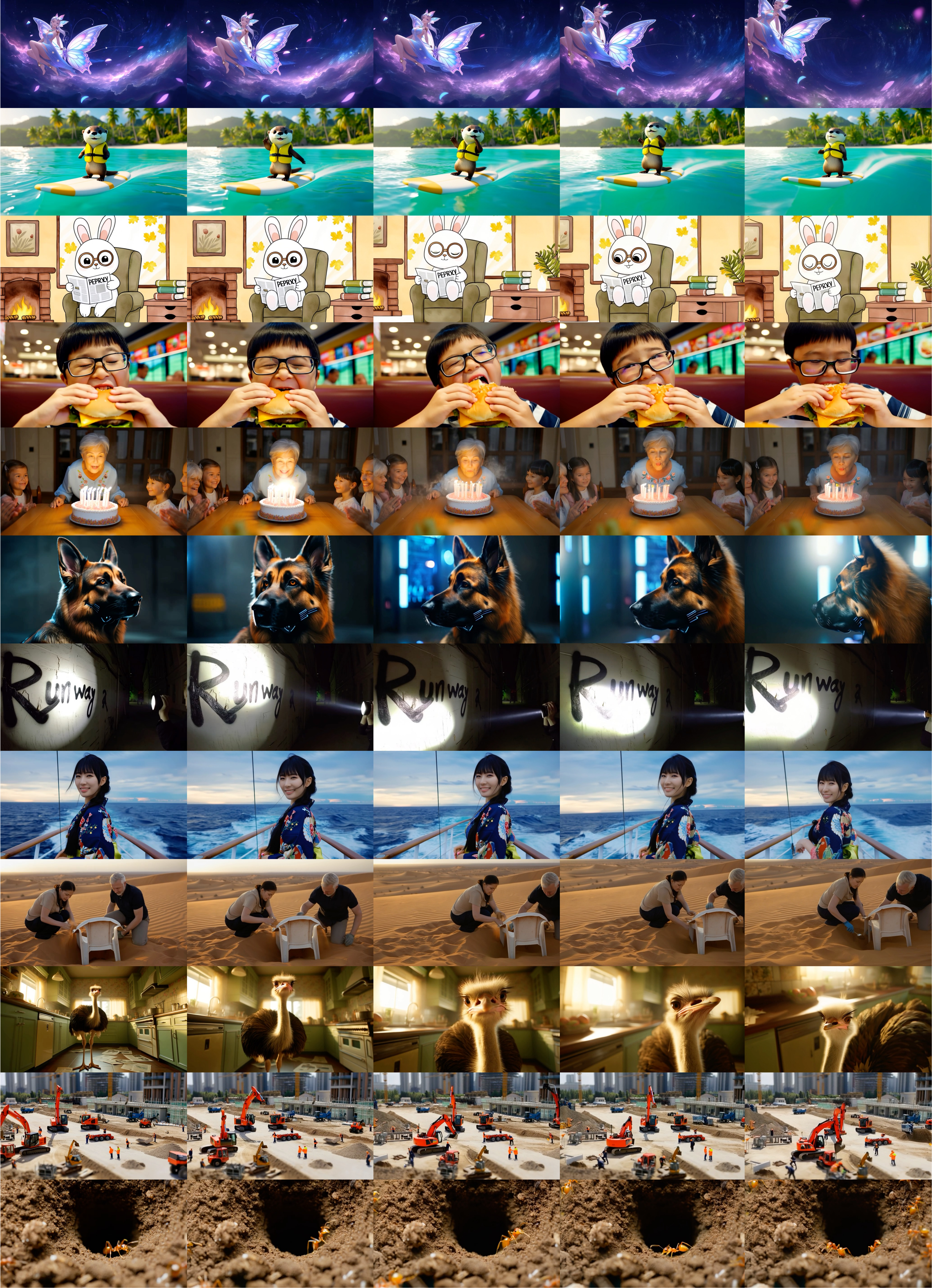}
    \vspace{-0.2in}
    \caption{\textbf{Additional qualitative results of 30s long video generation. }
    Our model produces temporally coherent and visually consistent sequences across diverse scenarios, further demonstrating its robustness.}
    \label{fig:addvis}
    \vspace{-0.1in}
\end{figure*}

To better demonstrate the robustness of our model, we present additional experimental results on 30s long video generation, as shown in Figure~\ref{fig:addvis}.

\end{document}